\documentclass[lettersize,journal]{IEEEtran}
\usepackage{amsmath,amsfonts}
\usepackage{algorithmic}
\usepackage{algorithm}
\usepackage{array}
\usepackage[caption=false,font=normalsize,labelfont=sf,textfont=sf]{subfig}
\usepackage{textcomp}
\usepackage{stfloats}
\usepackage{url}
\usepackage{verbatim}
\usepackage{graphicx}
\usepackage{cite}
\usepackage{ragged2e} 
\usepackage{booktabs,makecell, multirow, tabularx}
\usepackage{amssymb}
\usepackage{color}
\usepackage[mathscr]{euscript}
\begin{document}

\title{Fourier Boundary Features Network with Wider Catchers for Glass Segmentation}

\author{Xiaolin Qin,~\IEEEmembership{Senior Member,~IEEE,} Jiacen Liu, Qianlei Wang, Shaolin Zhang, Fei Zhu, Zhang Yi,~\IEEEmembership{Fellow,~IEEE} 
\thanks{Xiaolin Qin, Jiacen Liu, Qianlei Wang, and Shaolin Zhang are with the Chengdu Institute of Computer Applications at Chinese Academy of Sciences, Chengdu, 610213, China.Email: qinxl2001@126.com, dreakliu123552@163.com, wangqianlei36@gmail.com, z84489916@gmail.com}
\thanks{Xiaolin Qin, Qianlei Wang, and Shaolin Zhang are also with the School of Computer Science and Technology at University of Chinese Academy of Sciences, Beijing, 101408, China.}
\thanks{Fei Zhu is with the Centre for Artificial Intelligence and Robotics at Hong Kong Institute of Science \& Innovation, Chinese Academy of Sciences, Hong Kong, 999077, China. Email: zhufei2018@ia.ac.cn.}
\thanks{Zhang Yi is with the School of Computer Science at Sichuan University, Chengdu, 610065, China. Email:  zhangyi@scu.edu.cn.}
\thanks{Xiaolin Qin, Qianlei Wang, and Zhang Yi are the corresponding authors.}
}

\markboth{Journal of \LaTeX\ Class Files,~Vol.~14, No.~8, August~2021}%
{Shell \MakeLowercase{\textit{et al.}}: A Sample Article Using IEEEtran.cls for IEEE Journals}

\maketitle

\begin{abstract}
Glass largely blurs the boundary between the real world and the reflection. The special transmittance and reflectance quality have confused the semantic tasks related to machine vision. Therefore, how to clear the boundary built by glass, and avoid over-capturing features as false positive information in deep structure, matters for constraining the segmentation of reflection surface and penetrating glass. We proposed the Fourier Boundary Features Network with Wider Catchers (FBWC), which might be the first attempt to utilize sufficiently wide horizontal shallow branches without vertical deepening for guiding the fine granularity segmentation boundary through primary glass semantic information. Specifically, we designed the Wider Coarse-Catchers (WCC) for anchoring large area segmentation and reducing excessive extraction from a structural perspective. We embed fine-grained features by Cross Transpose Attention (CTA), which is introduced to avoid the incomplete area within the boundary caused by reflection noise. For excavating glass features and balancing high-low layers context, a learnable Fourier Convolution Controller (FCC) is proposed to regulate information integration robustly. The proposed method has been validated on three different public glass segmentation datasets. Experimental results reveal that the proposed method yields better segmentation performance compared with the state-of-the-art (SOTA) methods in glass image segmentation. 
\end{abstract}


\section{Introduction}
\IEEEPARstart{A}{s} a solid, glass has special properties of both light transmission and reflection, hence it is widely used in many scenes, such as streets, buildings, and industrial production workshops. At the same time, due to prior knowledge and the complex visual nervous system, humans can detect the glass, what is behind the glass, and the reflected virtual world, which is a huge task for machines. To reduce the impact of glass on the practical tasks of intelligent systems based on machine vision (e.g., depth estimation and driving obstacle avoidance), machine vision is required to accurately segment the glass region.

Previous exploration of glass semantic segmentation consistently found boundary detection and context information as two effective solutions. In some complex scenes, the glass only provides weak or incomplete boundaries, and the context content is highly confused by the glass transmission and reflection, resulting in reduced segmentation accuracy of networks. Although the advance of current methods to counter
\begin{figure}[tbp]
	\begin{center}
		\includegraphics[width=1\linewidth]{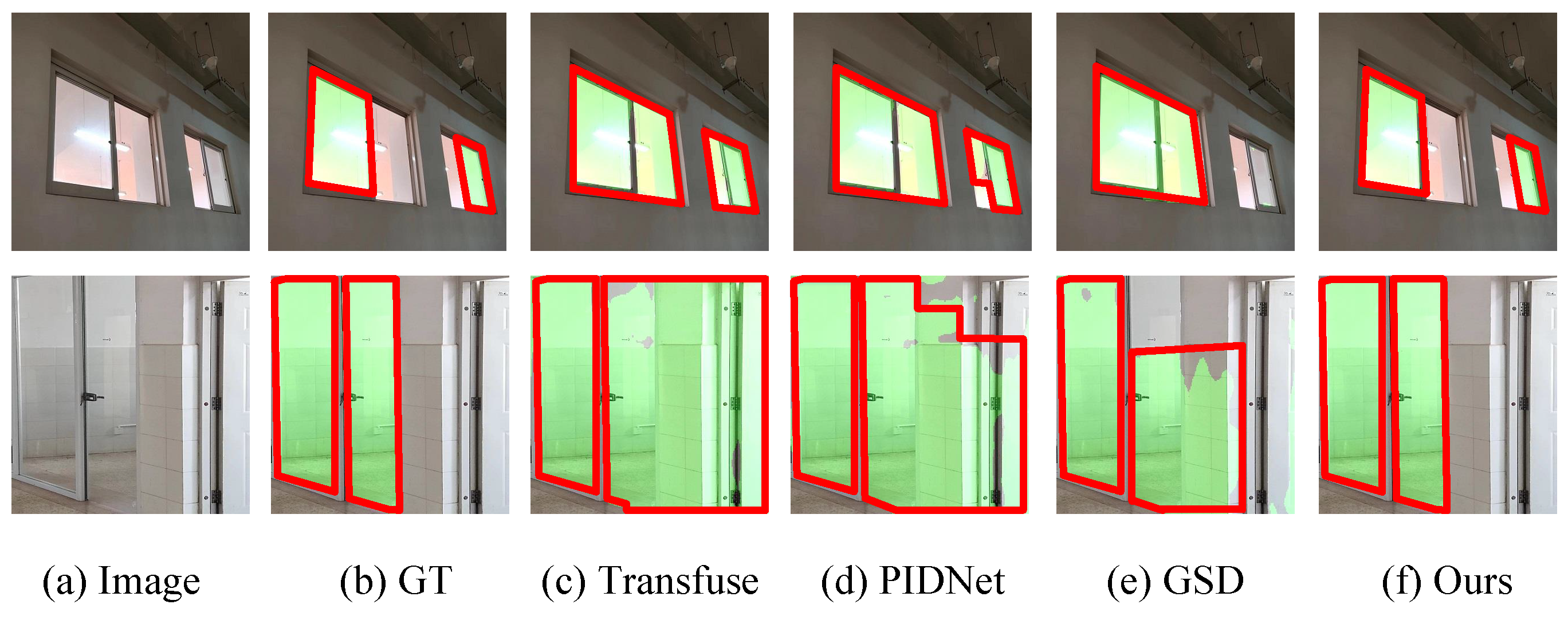}
	\end{center}
	\vspace{-12pt}
	\caption{Visual examples of glass segmentation. Compared with the SOTA methods, it can be concluded that the proposed network can accurately locate the segmentation boundary through a shallow feature capture framework and boundary constraints, and at the same time, the segmentation region is restricted within the boundary to avoid over-capturing and maintain regional consistency.}
	\label{fig1}
\end{figure}
the ambiguous glass boundaries and incoherent contextual information, there are absence of excavating the essential challenges: (1) Overabundant semantic information: the deep complex backbone model can extract fine-grained information to enhance the segmentation object detail. However, over-capturing essentially mismatches the glass segmentation because light transmission causes a strong context consistency between internal and external segmented areas, which is not susceptible to the affinity of highly extracted features; (2) Shaky boundary constraints: current methods utilize various boundary detectors to preserve the outline and locate regions, such as the shallow-low features, which are fragile and easily affected by the subsequent gradual abstraction and fine-granularity of the network.

To alleviate the issues above, we designed the Fourier Boundary Features Network with Wider Catchers (FBWC), which is a shallow-width glass segmentation network with stable boundary constraints. First, we introduce a novel feature extraction backbone to solve the critical problem of semantic information over-capturing. Specifically, Wider Coarse-Catchers (WCC) consist of multiple Capturing Units (CUs) to provide earlier trough points in the encoding process than the traditional backbone. The structure of shallow and wide with connections of continuous CUs benefits glass coarse-grained region information extraction through a series of appropriate horizontal trough points. At the same time, we fully take advantage of decoding to embed strong and steady boundary constraint information, which is concentrated significantly throughout the backbone without dispersiveness and distortion in highly abstracted processes. For another characteristic of glass, we designed Cross Transpose Attention (CTA) specifically to resist the impact of noise reflected by glass on the shallow layers through the focusing object area. Based on large and coarse-grained regions, progressively strengthened boundary constraints, and features complemented with high focusing, the Fourier Convolution Controller (FCC) is pointedly proposed to consolidate shallow boundaries with Fourier transform and flexibly balance that heterogeneous information by self-learning convolution paradigm. To sum up, our contributions are as follows:

\begin{itemize}
\item To the best knowledge, we might be the first attempt to design a wide and shallow backbone to combat over-capturing in the glass segmentation task. Besides, continuous, stable, and strong boundary constraints are fully leveraged throughout our WCC. For an extra supplement, Cross Transpose Attention is proposed to focus on details for keeping regional integrity and consistency.

\item A learnable solution for utilizing the heterogeneous features to self-adjust and balance the fusion is innovatively provided,  which provide an optimal combination of coarse-grained and fine-grained feature representations by embedding classical convolution theory. Meanwhile, the Fast Fourier Transform (FFT) is embeded to enhance the generated boundary features.

\item Experimental results on three different public glass segmentation datasets show that the proposed FBWC outperforms previous SOTA methods. We further demonstrate the ability of our network to reveal the nature of segmentation tasks based on glass properties.
\end{itemize}

\section{ Related Works}
\subsection{Boundary Detection}
From the perspective of bionics, one of the important bases for humans to judge glass-like objects is to find their boundary \cite{8}, which is a significant clue to help us divide reality and the virtual world transmitted and reflected by glass. Hence, the existing methods \cite{5,6,7} have carried out a lot of work around the glass boundary detection and made great progress. Xie \textit{et al.} \cite{1} found that due to high contrast in the edge of the glass, boundary detection is prone to observe compared with content for being paired with human visual perception. Therefore, they proposed a novel network named TransLab to utilize the predicted boundary map for improving accuracy. He \textit{et al.} \cite{2} enhanced boundary learning for glass segmentation through the refined differential module and further edge-aware point-based graph convolution to establish a complete shape along the boundary. At the same time, Cao \textit{et al.} \cite{3} argue that boundaries with large smaller amounts, the boundary-related imbalance problem, adversely affect glass semantic performance, which is solved by designed FakeMix. However, the mentioned methods of glass boundary detection heavily rely on the quality or quantity of boundary features extracted and optimized by the network, which is a worrying process due to weakening, distortion, and even dispersion of boundary during progressively abstracting. For this, Li \textit{et al.} \cite{4} optimized edge pixels by explicitly sampling under decoupled supervision for better consistency. Similarly, indirect boundary supervision through a single pathway is shaky and vulnerable to the erosion of high-dimensional semantic features. Inspired by previous works, we fully utilize boundary constraints directly with multiple supervision, aiming at offering firm and robust constraint features in the decoding process.

\subsection{Context Features of Glass}
The instability of boundary constraints and weakly blurred boundaries provided by glass in some complex scenes \cite{9} limit the favorable influence of glass boundary detection methods on the accuracy of segmentation results. Therefore, the methods are gradually biased towards the study of how to efficiently leverage abundant context for glass segmentation from multi-scale features extraction focusing \cite{7}\cite{16,17} and effective fusion \cite{18,19}. Yu \textit{et al.} \cite{10} have considered deep-layer features with more high-level semantics to be better at the location. Shallow layer features own larger spatial sizes with detailed low-level information instead. PGSNet is designed through progressively aggregating to avoid fusing these features naively. Hu \textit{et al.} \cite{14} proposed a VIT-based deep architecture to associate multi-layer receptive field features and optimize fusion results. Besides, due to the large and complete glass regions context, the attention mechanism is in full utilization. CAGNet \cite{11} integrates two kinds of convolutional attention mechanisms with a Transformer head for deep-shallow feature analysis and fusion. Hou \textit{et al.} \cite{12} further introduce the Transformer encoder-decoder architecture to boost the usage of multi-levels of context information. Lin \textit{et al.} \cite{13} embed context correlation into the attention module to bridge contextual links among objects both spatially and semantically.

Although complex and deep network structures can abstract spatial semantic features on multi-scales to facilitate subsequent feature fusion, for the special segmentation target, glass, over-captured segmentation features will destroy the coherence of shallow information, making it difficult for the network to maintain the integrity of large-area segmentation. At the same time, the previous attention mechanism is more inclined to deal with multi-scale features with different depths and shallow features, and it fails to effectively match the width feature extraction framework to complete the shallow feature self-focusing task.
\begin{figure*}
	\begin{center}
		\includegraphics[width = 1\linewidth]{figure/glass.pdf}
	\end{center}
	\caption{Architecture of the proposed FBWC. Given an input glass image, there will be two branches to receive the same picture and change them into different features. In particular, one of features is fed into WCC to alleviate the over-capturing and boundary constraints of scarcity. The others is processed by the CTA to supplement and focus fine-grained information. The designed FCC as a learnable feature fusion regulator to balance heterogeneous inputs. Finally, a fine-grained weighting factor regulating the shallow and large area segmentation result with supplementing boundary information constraints to achieve the purpose of dynamic optimized network predictions. The number next to each module denotes its parameter. Express model concerns more intuitively by exporting Hot Map.}
	\label{fig2}
\end{figure*}

\subsection{Architecture of networks}
In the classical network structure, deeper layers benefit the network's ability to abstraction, which is conducive to the affinity for high-level object features and captures rich and complex semantic information. In addition to the problem of gradient disappearance or explosion \cite{28} with deepening architecture, depth-based networks often fall into the dilemma of over-capturing for the task of semantic segmentation of glass. Specifically, most of the glass-like objects are a large area with a complete boundary and regular shape, hence over-abstraction of this information is harmful to improving the performance of the segmentation. At the same time, the shallow features learning in the network can also learn the complex functions and information captured by the deep net \cite{20}. Therefore, in order to better pair the object's characteristics of segmentation, a suitable network architecture with carefully balanced depth and width \cite{21} is proposed to avoid the accuracy decrease caused by over-segmentation. Embracing the advantages of shallow layers, the basic low-level shape and outline of the glass can be relatively intact without distortion. Aiming at the reservation of shallow features, trough points, as turning points in the depth of the network, should occur in advance to avoid the dilution effect \cite{22} of features during deepening. Inspired by the Broad Learning System \cite{24,25,26,27}, multiple trough points generated by the recurrent connections structure \cite{23} widen the structure and drive the extracted information gradually permeating the network.

To mitigate the limitations associated with the conventional network framework in glass segmentation, such as overfitting from deep networks and the loss of intricate features from shallow networks, we propose a cyclic shallow backbone network. This network imposes rigorous constraints on layer-by-layer boundaries, enabling the extraction of semantic information tailored to the unique characteristics of the glass region. Thus, we attempt to explore the wider feature extraction to avoid over-capturing caused by a deeper network, which might be not suitable for the special semantic objects with large areas. This approach safeguards feature integrity and enhances segmentation accuracy by ensuring that semantic information is effectively captured within the boundary range.
\section{Proposed Methodology}
The glass can be easily recognized by the human visual system based on prior knowledge, even though eyes suffer from the noise of objects through the glass and what the glass reflects. Humans are prone to utilize the surrounding context to confirm the existence of glass boundaries. With the information about the boundary, humans can accurately distinguish between the real world and the virtual world reflected by glass. This is particularly important for how machine vision circumvented the confusion of information brought by the properties of glass, both transmission and reflection. Inspired by this, we attempted to encode this practical phenomenon to achieve the bionic human visual system's intention to segment glass regions, then we proposed the FBWC. 
\subsection{Architecture}
The overall architecture of the proposed FBWC is introduced in Figure \ref{fig2}. Firstly, for the input $x \in \mathbb{R}^{3 \times H \times W}$,we utilize downsampling and convolution operation as pretreatment to preliminarily obtain the same two features: $x_{t r i} \in \mathbb{R}^{C \times \frac{H}{\lambda} \times \frac{W}{\lambda}}$ and $x_{c i r} \in \mathbb{R}^{C \times \frac{H}{\lambda} \times \frac{W}{\lambda}}$ where $C$ denotes the number of channels and $\lambda$  downsampled shape hyperparameter.

For $x_{c i r}$, we introduce WCC, which consists of $I$ CUs, to acquire large and shallow information through lateral broadening backbone without longitudinal deepening. Besides, the glass edge information extracted by the canny operator from ground truth is exploited to constrain the boundary in WCC. Then, CTA is for $x_{tri}$ elaborated, which combines effective mechanism of attention between two feature branches to supplement and focus detailed information. Moreover, the core feature fusion regulator module of FBWC is designed as a FCC with a learnable paradigm to flexibly balance deep-shallow heterogeneous feature expression.

Based on the self-learning fusion feature controller, the FBWC derives a lot of adjustment flexibility and self-correction margin. Thus, instead of designing a complex loss function design, we introduce ohem cross-entropy \cite{shrivastava2016training} and binary cross-entropy loss to monitor the propagation of semantic information across the proposed network. To be specific, for the $i$-th decoder of CU in WCC, binary cross-entropy loss with boundary-awareness is applied by the Canny operator to gradually strengthen boundary constraints. At the same time, $\ell_{Seg}$ and $\ell_{AM}$ refer to supervised learning from ground truth to the controller and Attention Mechanism (AM) results with ohem cross-entropy loss ,  respectively. Therefore, the final loss for FBWC is:
\begin{equation}
	L o s s=\ell_{Seg}+\ell_{AM}+\frac{1}{{N_{CUs}}} \sum_{{\overline{i}}=1}^{N_{CUs}} \ell_{\overline{i}} ,
\end{equation}
where ${N_{CUs}}$ represents numbers of CUs, and $\{\ell_{\overline{i}} | {\overline{i}}\in{N_{CUs}}\}$ denotes the loss of Boundary Constraint (BC Loss) to a single CU. 

\subsection{Wider Coarse-Catchers}
The point at both large segmentation and transmittance areas like the object of glass, we hold the opinion that over-capturing with classical deep structure embraces noise instead. Due to deep layers' highly abstracting semantic information, a backbone with a complex semantic concentration frame generates dispersive fine-grained features which make a barrier for boundary constraints and glass surface segmentation. Therefore, taking advantage of low-level semantic information in shallow layers is conducive to pairing the implicit characteristics of the glass.

\begin{figure}[tbp]
	\begin{center}
		\includegraphics[width=1\linewidth]{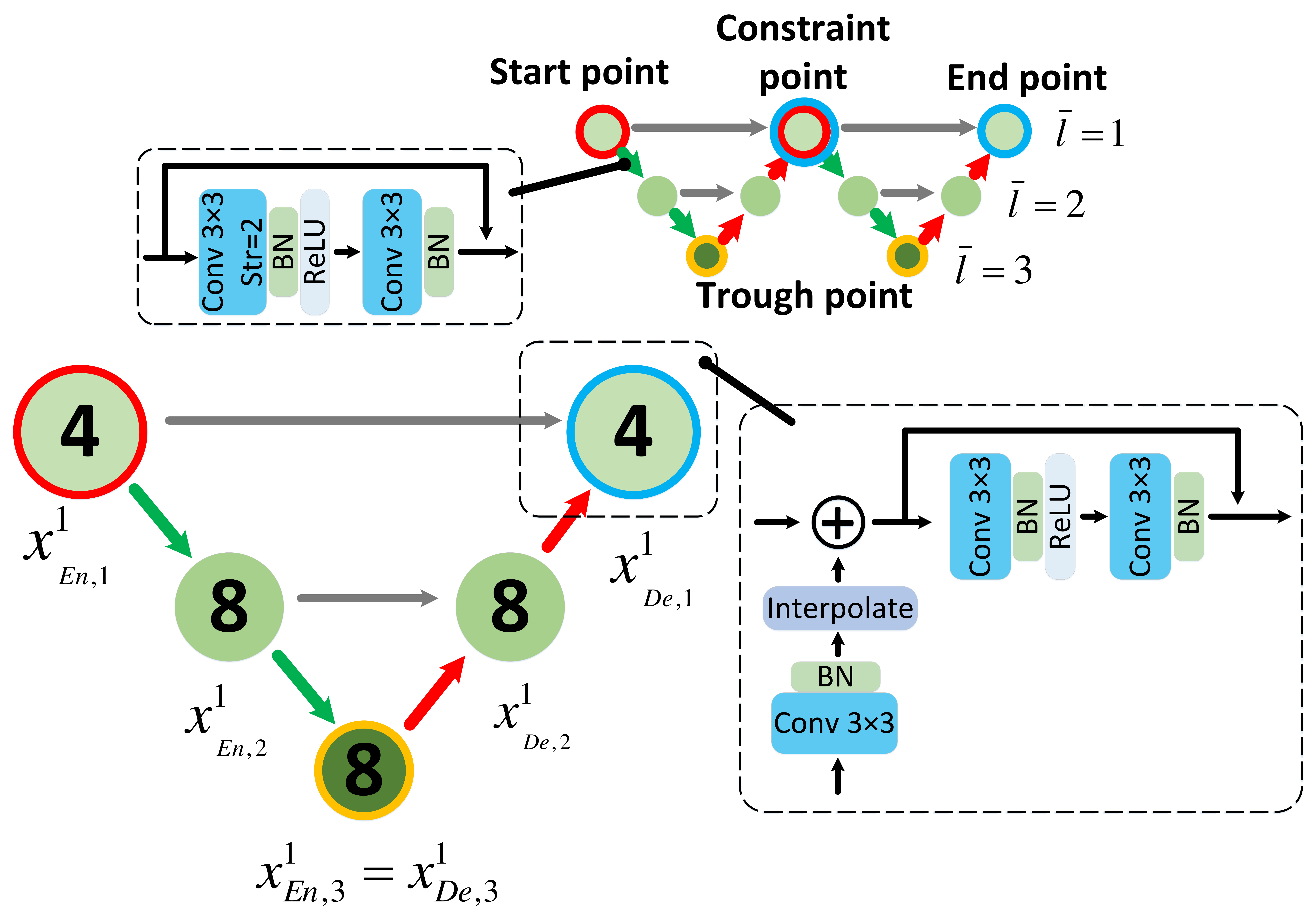}
	\end{center}
	\vspace{-12pt}
	\caption{Details of Capturing Unit on WCC, which is designed to avoid over-capturing and provide boundary constraint.}
	\label{fig3}
\end{figure}

As shown in Figure \ref{fig3}, we employ CUs as the component of our WCC. We notice that shallow encoded information can provide a large area of segmentation, and the decoding process is conducive to boundary constraints. Hence, the CU consists of three layers $\overline{l} \in[1,2,3]$ encoding-decoding symmetric structure and three important points: start point, trough point, and end point. In particular, the constraint point is named by sharing the both start and end position in WCC.

Specifically, in the encoding process, we use convolution and down-sampling to manage the number of channels and half the size of height and width respectively on each layer, which can be formulated as:
\begin{equation}
x_{E n, \overline{l}}^{\overline{i}}=\Gamma^{\overline{l}-1}\left\{\omega \cdot \vartheta^{3 \times 3}\left(x_{E n, 1}^{\overline{i}}\right)\right\},\overline{l} \in[2,3],
\end{equation}
where $\omega$ and $\vartheta^{3 \times 3}$ denote dowm-sampling and convolutional operator,  respectively. And the $\Gamma^{{\overline{l}}}\{\cdot\}$ is defined as the continuous operator which executes ${\overline{l}}$ times. 

For the ${\overline{i}}$-th CU, $x_{E n, 3}^{\overline{i}}$ or $x_{De, 3}^{\overline{i}}$ and $x_{E n, 1}^{\overline{i}}$ are named trough point and start point,  respectively. After the continuous operation of convolution and dowmsampling, in the process of decoding, we apply up-sampling operators with bilinear interpolation $\mu$ and the skip connection to bridge result features of up-sampling and $x_{E n, {\overline{l}}}^{\overline{i}}$. Finally, the decoding process is defined as:
\begin{equation}
x_{D e, {\overline{l}}}^{\overline{i}}=\vartheta^{3 \times 3}\left[Skip\left(\mu\left(x_{D e, {\overline{l}}+1}^{\overline{i}}\right), x_{E n, {\overline{l}}}^{\overline{i}}\right)\right],{\overline{l}} \in[1,2],
\end{equation}
where $x_{D e, {\overline{l}}}^{\overline{i}}$ denotes the features of the ${\overline{l}}$-th layer on the decoder of ${\overline{i}}$-th CU. And the operator of skip connection $Skip(\cdot, \cdot)$ is defined as $Skip(a, b)=a \oplus b$.
To sum up, the CU mapping ( $\Xi_{CU}$ ) can be expressed as:
\begin{equation}
\Xi_{CU}(z)=Skip(\check{z}, \tilde{z}),
\end{equation}
where ${\check{z}}$ represents continuously downsampling via convolution. Furthermore, $\tilde{z}$ is denoted as upsampling the results of ${\check{z}}$. The process of input $x_{\text {cir }}$ getting outputs after the first CU is formulated as:
\begin{equation}
\left[\begin{array}{l}
x_{D e, 1}^1 \\
x_{E n, 3}^1
\end{array}\right]=\Xi_{CU}\left(x_{c i r}\right).
\end{equation}
where $x_{E n, 3}^1$ is for coarse-grained features acquisition and $x_{D e, 1}^1$ is for continuous asymptotic boundary constraints. Generally, when ${\overline{i}} \in[1,2, \ldots, N]$, it means that multiple CUs need to be connected in series. As a result, the representation of variables on output layers in CUs are defined as following:
\begin{equation}
\left\{\begin{array}{l}
x_{E n, 1}^1=\omega \cdot \vartheta^{3 \times 3}\left(x_{c i r}\right), {\overline{i}}=1 \\
x_{E n, 1}^{\overline{i}}=x_{D e, 1}^{{\overline{i}}-1}, {\overline{i}} \neq 1. \\
\left[\begin{array}{l}
	x_{D e, 1}^{\overline{i}} \\
	x_{E n, 3}^{\overline{i}}
\end{array}\right]=\Gamma^{\overline{i}}\left\{\Xi_{CU}\left(x_{c i r}\right)\right\} , others
\end{array}\right.
\end{equation}

Finally, the constraint points set $X_{D e, 1}=\left\{x_{D e, 1}^{\overline{i}}\right\}_{{\overline{i}}=1}^N$ and trough points set $X_{E n, 3}=\left\{x_{E n, 3}^{\overline{i}}\right\}_{{\overline{i}}=1}^N$ are represented as the outputs of multiple CUs for forming a wider shallow feature capture framework.

\subsection{Cross Transpose Attention}
Through WCC as backbone, trough point and constraint point gather large glass semantic region and strong constraint boundary information respectively to avoid over-capturing noise caused by the structure of trough point lagging. However, shallow and wide networks with early troughs are prone to suffer from unfocused features, which
\begin{figure}[tbp]
	\begin{center}
		\includegraphics[width=1\linewidth]{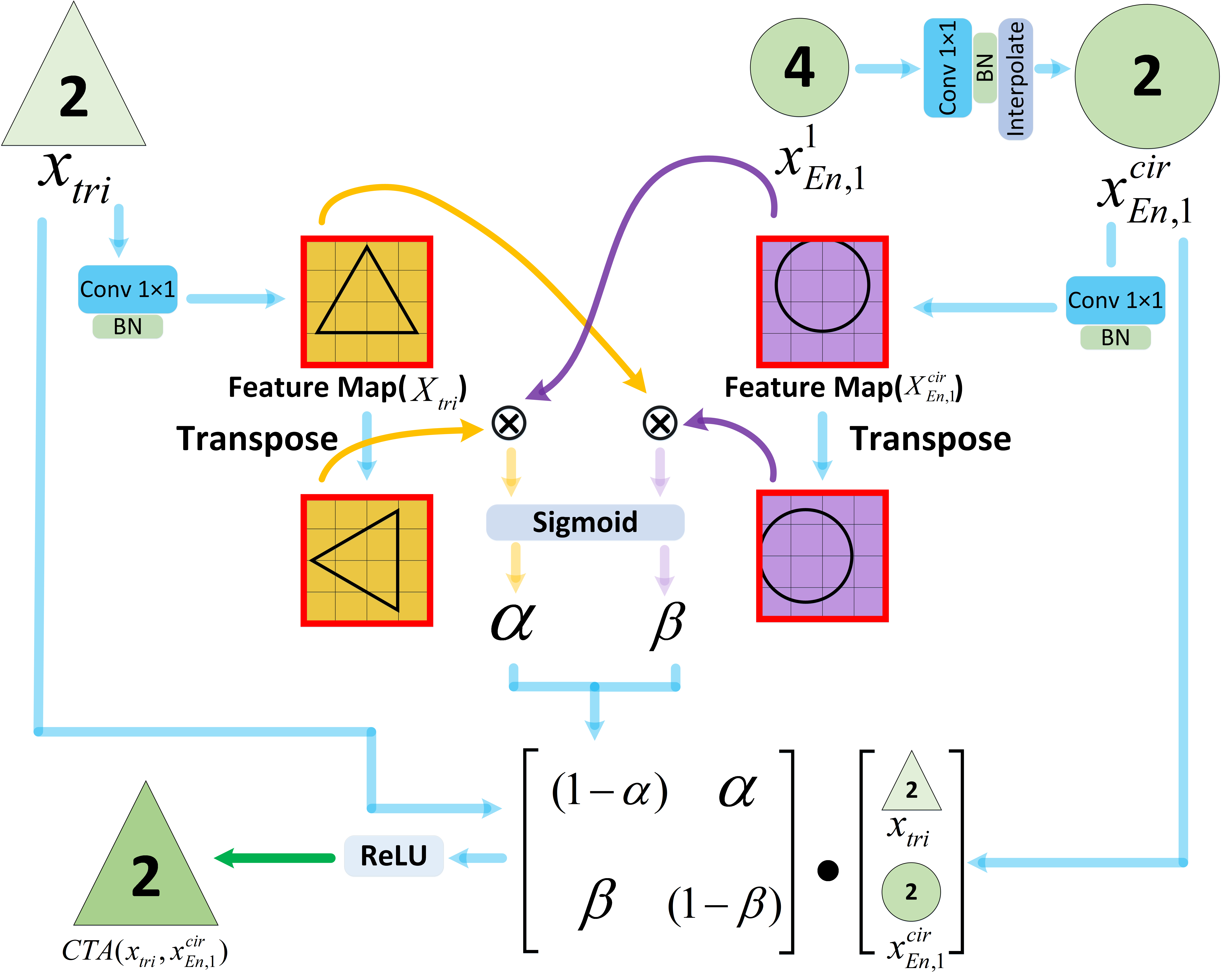}
	\end{center}
	\vspace{-12pt}
	\caption{Illustration of Cross Transpose Attention (CTA) blocks.}
	\label{fig4}
\end{figure}
results in incomplete semantic regions within the boundary due to glass reflection. To supplement the fine-grained features and keep regional consistency, CTA is designed as an auxiliary branch for accurate segmentation, which is illustrated in Figure \ref{fig4}. 

For the input $x_{t r i}$, $x_{E n, 1}^1$ and $x_{E n, 2}^1$, we first keep the image size consistent through bilinear interpolation for obtaining $x_{E n, 1}^{c i r}$ and $x_{E n, 2}^{c i r}$. As for $x_{t r i}$ and $x_{E n, 1}^1$, mapping the same size tensors separately utilizing convolution: $X_{tri}=BN(\vartheta^{1 \times 1}\left(x_{t r i}\right))$ and $X_{E n, 1}^{c i r}=BN(\vartheta^{1 \times 1}\left(x_{E n, 1}^{c i r}\right))$. Then, we transpose the respective tensors to cross multiply at the same pixel position for calculating the weight of the pixel. By the generated cross attention weights, a joint matrix is established to obtain updated attention results. We formulate the mentioned process:
\begin{equation}
\alpha=\sigma\left(X_{E n, 1}^{c i r} \cdot X_{t r i}{ }^T\right),
\end{equation}
\begin{equation}
\beta=\sigma\left(X_{t r i} \cdot {X_{E n, 1}^{cir }}^T\right),
\end{equation}
\begin{equation}
C T A\left(x_{t r i}, x_{E n, 1}^{c i r}\right)=\left[\begin{array}{lr}
(1-\alpha) & \alpha \\
\beta & (1-\beta)
\end{array}\right] \cdot \left[\begin{array}{l}
x_{t r i} \\
x_{E n, 1}^{c i r}
\end{array}\right],
\end{equation}where $\sigma$ refers to Sigmoid function and $C T A(a, b)$ is denoted as Cross Transpose Attention operator with tensors $a$ and $b$. The cross similarity coefficients are expressed by $\alpha$ and $\beta$. The final output of CTA is $x_{t r i}^{\text {out }}=C T A\left(\left(C T A\left(x_{t r i}, x_{E n, 1}^{c i r}\right), x_{E n, 2}^{c i r}\right)\right)$ for effective focusing on fine-grained semantic features.

\subsection{Fourier Convolution Controller}
\begin{figure*}
	\begin{center}
		\includegraphics[width = 1\linewidth]{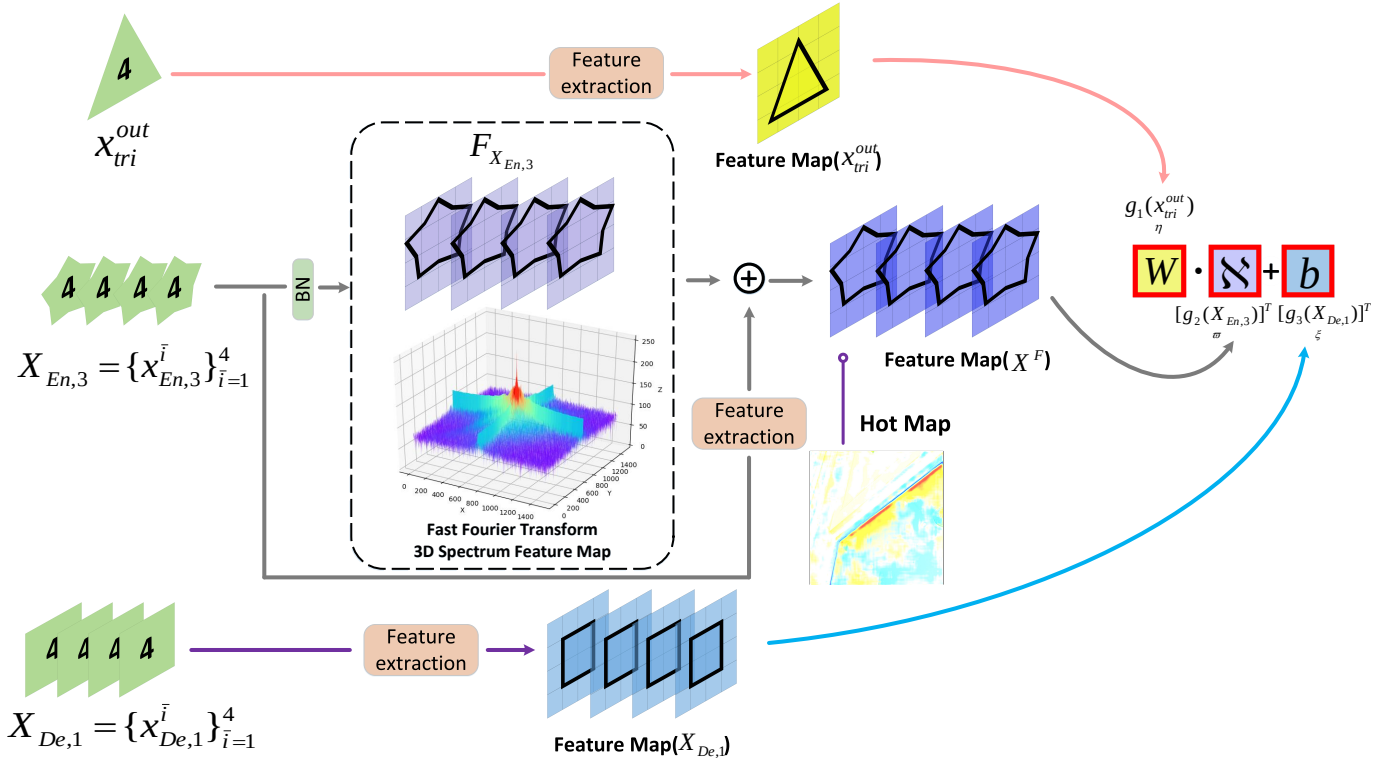}
	\end{center}
	\caption{Illustration of FCC, which serves as the core component of FBWC for providing a flexible solution to the features fusion problem. Express model concerns more intuitively by exporting Hot Map.}
	\label{fig5}
\end{figure*}

Instead of directly fusing information among $x_{t r i}^{\text {out }}$, $X_{E n, 3}$, and $X_{D e, 1}$, the FBWC utilizes FCC as the features learnable regulator to deal with heterogeneous inputs. Figure \ref{fig5} illustrates the detailed structure of our FCC, which is designed to provide an optimal combination of coarse-grained and fine-grained feature representations by embedding classical convolution theory. This feature combination is a nonlinear combination of features under different branches:
\begin{equation}
	\underset{\eta, \varpi, \xi}{\arg \min }\left(Y-\underset{\eta, \varpi, \xi}{f}\left(\underset{\eta}{g_1}\left(x_{\text {tri }}^{\text {out }}\right), \underset{\varpi}{g_2}\left(X_{En,3}\right), \underset{\xi}{g_3}\left(X_{De,1}\right)\right)\right),
\end{equation}where $\underset{\eta}{g_1(\cdot)}, \underset{\varpi}{g_2}(\cdot), \underset{\xi}{g_3(\cdot)}$ represent the feature extraction of three different branches under parameter sets $(\eta, \varpi, \xi)$, the $ f(\cdot, \cdot, \cdot)$ means model training and output the feature after fusion, and the parameter sets $(\eta, \varpi, \xi)$ are learnable. The $Y$ represents the desired feature which determined by ground-truth. 

We use $X^F$  to represent the result of $\underset{\varpi}{g_2}\left(X_{En,3}\right)$ which fused the glass frequency domain information obtained through FFT:
\begin{equation}
	X^F  =\vartheta^{1 \times 1}(X_{En,3}) + {\boxplus}_{x^{\overline{i}}_{En,3}} [F_{x_{En,3}}(c, d)], {\overline{i}} \in [1,2,3,4],
\end{equation}
where the ${\boxplus}_{x^{\overline{i}}_{En,3}} [ \cdot ]$ means executing  operation for each channel of $x^{\overline{i}}_{En,3}$ which belongs to $X_{En,3}$, and the $F_{x_{En,3}}(c, d)$ represents the glass frequency domain information extracting process:
\begin{equation}
\begin{aligned}
   F_{x_{En,3}}(c, d) =\sum_{w^*=0}^{\frac{W}{\lambda}-1} \sum_{h^*=0}^{\frac{H}{\lambda}-1} x_{En,3}(w^*, h^*) e^{-j 2 \pi\left(\frac{\lambda c w^*}{W}+\frac{\lambda d h^*}{H}\right)},
\end{aligned}
\end{equation}where $c \in[0, \ \dots \ ,\frac{W}{\lambda}-1]$ and $d \in[0, \ \dots \ ,\frac{H}{\lambda}-1]$ refer to frequency domain variables. In the FFT process, the $x_{En,3}(w^*, h^*)$ is the value of the feature point at position $(w^*, h^*)$ in one channel belongs to $x^{\overline{i}}_{En,3}$.

The two-dimensional FFT can be regarded as a generalization of one-dimensional FFT, then we obtain expression of a single feature point value on a single column:
\begin{equation}
	x_{En,3}( \cdot, h^*)=\frac{a_0}{2}+\sum_{n=1}^P\left[a_n \cdot \cos (n w h^*)+b_n \cdot \sin (n w h^*)\right].
\end{equation}

We replace $\cos (n w h^*)$ and $\sin (n w h^*)$ by $\left(e^{i n w h^*}+e^{-i n w h^*}\right) / 2$ and $-i\left(e^{i n w h^*}-e^{-i n w h^*}\right) / 2$ respectively, then Formula (13) can be reformed as:
\begin{equation}
\begin{aligned}
	&x_{En,3}(\cdot, h^*)=\frac{a_0}{2} \\
	&+\sum_{n=1}^P\left[\frac{a_n-i b_n}{2} \cdot e^{i n w h^*}+\frac{a_n+i b_n}{2} \cdot e^{-i n w h^*}\right].
\end{aligned}
\end{equation}

Let $c_0=\frac{a_0}{2}$, $c_n=\frac{a_n-i b_n}{2}$, $c_{2 P+1-n}=\frac{a_n+i b_n}{2}$, where $N=2 P+1=\frac{H}{\lambda}-1$, $n \in [1,2,3,...,P]$, then we obtain a more unified form:
\begin{equation}
	\begin{aligned}
	x_{En,3}(\cdot, h^*)
	&=c_0 \cdot e^{i h^* w \cdot 0}+c_1 \cdot e^{i h^* w \cdot 1}+c_2 \cdot e^{i h^* w \cdot 2}+ \ \dots \ \\
	&+c_{N-1} \cdot e^{i h^* w \cdot N-1}, w=2 \pi / N.
	\end{aligned}
\end{equation}
The process of FFT can be thought of as the solution of the inverse of in the following equation:
\begin{equation}
	\Theta \cdot \left[\begin{array}{c}
		c_0 \\
		c_1 \\
		c_2 \\
		\vdots \\
		c_{N-1}
	\end{array}\right]=\left[\begin{array}{c}
		x_{En,3}(\cdot, 0) \\
		x_{En,3}(\cdot, 1) \\
		x_{En,3}(\cdot, 2) \\
		\vdots \\
		x_{En,3}(\cdot, N-1)
	\end{array}\right],
\end{equation}where $\Theta=$
\begin{equation}
		\left[\begin{array}{ccccc}
		1 & 1 & \cdots & 1 \\
		e^{i w \cdot 0} & e^{i w \cdot 1} & \cdots & e^{i w \cdot(N-1)} \\
		e^{i 2 w \cdot 0} & e^{i 2 w \cdot 1} & \cdots & e^{i 2 w \cdot(N-1)} \\
		\vdots & \vdots & \ddots & \vdots \\
		e^{i(N-1) w \cdot 0} & e^{i(N-1) w \cdot 1} & \cdots & e^{i(N-1) w \cdot(N-1)}
	\end{array}\right]_{N \cdot N}.
\end{equation}

Then we use $\Theta^{-1}(j, k)$ to represnet the element in $j$-th row and $k$-th column in $\Theta^{-1}$, then $c_j$ can be expressed as:
\begin{equation}
	c_j=\sum_{k=0}^{N-1} x_{En,3}(k) \cdot \Theta^{-1}(j, k), j \in [0,1,...,N-1].
\end{equation}

Let $j=j_{m-1} \cdot 2^{m-1}+ \ \dots \ +j_v \cdot 2^v+ \ \dots \ +j_1 \cdot 2+j_0$, $k=k_{m-1} \cdot 2^{m-1}+ \ \dots \ +k_v \cdot 2^v+ \ \dots \ +k_1 \cdot 2+k_0$, where $j_v=0$ or $1$ and $k_v=0$ or $1$, and  we define operation $\left[j_m, \ \dots \ , j_1, j_0\right]^*$ as follow:
\begin{equation}
	\left[j_m, \ \dots \ , j_1, j_0\right]^*=j_m \cdot 2^m+ \ \dots \ +j_1 \cdot 2+j_0,
\end{equation}
Then we transfer Formula (19) into the following form:
\begin{equation}
	\begin{aligned}
	& c_{[j_{m-1}, \ \dots \ , j]^*} = \sum_{k_0} \sum_{k_1} \ \dots \ \sum_{k_{m-1}} x_{En,3}\left(\left[k_{m-1}, \ \dots \ , k_0\right]^*\right)\\ 
	&\cdot \Theta^{-1}\left(j, k_{m-1} \cdot 2^{m-1}\right) \cdot \Theta^{-1}\left(j, k_{m-2} \cdot 2^{m-2}\right) \\
	& \ \dots \  \Theta^{-1}\left(j, k_0\right),
	\end{aligned}
\end{equation}

When $\Theta^{-1}\left(j, k_{m-1} \cdot 2^{m-1}\right)=\Theta^{-1}\left(j_0, k_{m-1} \cdot 2^{m-1}\right)$, we can obatin the inner sum over $k_{m-1}$:
\begin{equation}
	\begin{aligned}
	& x_{En,3(l)}\left(\left[j_0, \ \dots \ , j_{l-1}, k_{m-l-1}, \ \dots \ , k_0\right]^*\right) \\
	& =\sum_{k_{m-1}} x_{En,3}\left(\left[k_{m-1}, \ \dots \ , k_0\right]^*\right) \cdot \Theta^{-1}\left(j_0, k_{m-1} \cdot 2^{m-1}\right),
	\end{aligned}
\end{equation}
where $l$ is the intercept location, if $l=1,2, \ldots, m$, and $m=\log _2 N$, we can get a more general form:		
\begin{equation}
	\begin{aligned}
		& x_{En,3(l)}\left(\left[j_0, \ \dots \ , j_{l-1}, k_{m-l-1}, \ \dots \ , k_0\right]^*\right)= \\
		& x_{En,3(l-1)}\left(\left[j_0, \ \dots \ , j_{l-2}, k_{m-l-2}, \ \dots \ , k_0\right]^*\right)+ \\ 
		&(-1)^{j_{l-1}} \cdot i^{j_{l-2}} x_{En,3(l-1)}\left(\left[\left[j_0, \cdots, j_{l-2}, k_{m-l-2}, \ \dots \ , k_0\right]^*\right]\right) \\
		&\cdot \Theta^{-1}\left(j_{l-3}, 2^{l-3} \cdot 2^{m-l}\right) \cdot \Theta^{-1}\left(j_{l-4}, 2^{l-4} \cdot 2^{m-l}\right) \\
		& \ \dots \  \Theta^{-1}\left(j_0, 2^{m-l}\right).
	\end{aligned}
\end{equation}

From Formula (11) and Formula (22), the FFT result is shown as follow:
\begin{equation}
	\begin{aligned}
		&X^F = \vartheta^{1 \times 1}(X_{En,3})+ {\boxplus}_{x^{\overline{i}}_{En,3}} [\sum_{w^*=0}^{\frac{W}{\lambda}-1} \sum_{l=1}^{\log _2 N} x_{En,3(l)}(w^*)] \\
		&,{\overline{i}} \in [1,2,3,4],l \in [1,2, \ldots, \log _2 N].
	\end{aligned}
\end{equation}

\begin{figure*}
	\begin{center}
		\includegraphics[width = 1\linewidth]{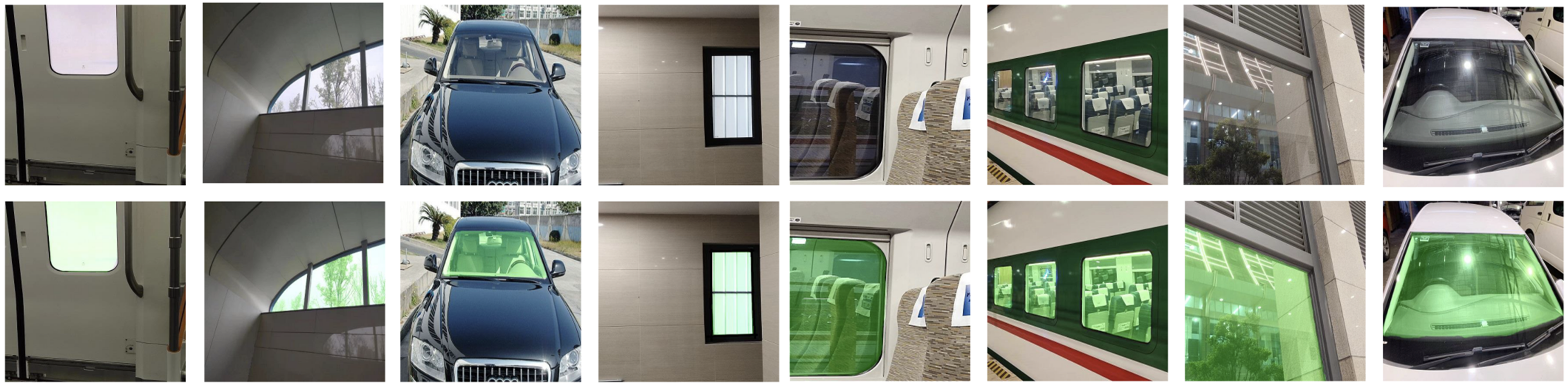}
	\end{center}
	\caption{More visual display of glass segmentation results of our FBWC beyound all the test sets.}
	\label{fig6}
\end{figure*}

According to Formula (22) and (23), there exists a recursive subtraction with a window size of 2 and a step size of 1 during the process of conducting FFT, which means that the difference between odd and even components is magnified in the frequency domain. Due to the images having naturally sparse representations \cite{shi2017new}, we only use the even symmetry components of each channel and the real part is reserved. According to \cite{zak2023fourier}, the IFFT is not required while the imaginary part is not used and the regular geometric structure feature in the pixel domain will be reserved. Then the boundary of glass in the pixel domain which is a regular geometric structure can be magnified in the frequency domain. However, using only frequency domain features may result in the loss of some refinement features in the pixel domain. To solve this problem, we fused the two heterogeneous source data in the frequency domain and spatial domain.

\textbf{Proposition 1.} \textit{Assuming $I^*$ is an image with regular geometric boundaries. For any continuous piecewise linear function subset $\Omega$ and any $f \in \Omega$, any convolutional neural network ( CNN ) $f_J^{\mathbf{w}, \mathbf{b}}$ with depth $J \in \mathbb{Z^+}$ and ReLU function exists:}
\begin{equation}
\begin{aligned}
&\left\|f(I^*)-f_J^{\mathbf{w}, \mathbf{b}}(FFT(I^*).real+I^*)\right\|_{\Omega} \\
&<\left\|f(I^*)-f_J^{\mathbf{w}, \mathbf{b}}(I^*)\right\|_{\Omega}.
\end{aligned}
\end{equation}

\textbf{\textit{Proof.}} For convenience, we consider the binary case, for simplicity, which is also suitable to the case in our study to explain how FFT affects CNN learning. Denote $D_0$ and $D_1$ to be the sets which contained the signals from two classes, and there is a positive gap between $D_0$ and $D_1$. That is, there exists some $d_0 > 0$ such as:
\begin{equation}
d\left(D_0, D_1\right)=\left\{\left\|x_0-x_1\right\| \mid x_0 \in D_0, x_1 \in D_1\right\}=d_0>0 .
\end{equation}
Assume that there is an ideal function $f$ such as:
\begin{equation}
f(x)=0, \quad \mbox { if } x \in D_0  \quad \mbox { and } \quad f(x)=1, \quad \mbox { if } x \in D_1 .
\end{equation}
For example, the $f(x)$ can be expressed as:
\begin{equation}
f(x)=
\begin{cases}
con_1 \cdot x+bias_1, & x \in D_0 \\ 
con_2 \cdot x+bias_2,  & x \in D_1 ,
\end{cases}
\end{equation}
where $con_1$ and $con_2$ represent the coefficients, at the same time, the $bias_1$ and $bias_2$ represent the biases. The $f$, which is the operation of the convolutional layer, is composed of multiplication and addition. Then we can apply a CNN consisting of at least one convolution layer to approximate Formula (27). To increase the approximating ability, we usually introduce a nonlinear activation function. Since FFT in our case will produce some unwanted negative values, we use ReLU as the nonlinear function. Further, assuming $D_0 \cup D_1 = I^*$, the $I^*$ represents an image with regular geometric boundaries. 

We input the image $I^*$ to a CNN with depth $J \in \mathbb{Z^+}$, and provided $S$ samples which are homologous to $I^*$ during the training. The approximating continuous piecewise linear function $f_J^{\mathbf{w^*}, \mathbf{b^*}}$ is obtained via solving the empirical minimization:
\begin{equation}
\min _{\mathbf{w},\mathbf{b}}
\frac{1}{S} \sum_{s=1}^S\left(f\left(I_s\right)-f_J^{\mathbf{w}, \mathbf{b}}(I_s)\right)^2.
\end{equation}

The approximating accuracy can be calculated as follows:
\begin{equation}
\left\|f(I^*)-f_{J}^{\mathbf{w^*}, \mathbf{b^*}}(I^*)\right\|
=\left\{\int_{I^*}\left(f(x)- f_{J}^{\mathbf{w^*}, \mathbf{b^*}}(x)\right)^2 d \mu\right\}^{1 / 2},
\end{equation}
where $\mu$ is a probability measure on $I^*$. Direct estimation of Formula (29) is difficult due to the additional complication of sampling errors, thus most existing approaches estimate Formula (29) by separating it from the approximation error via triangle inequality:
\begin{equation}
\begin{aligned}
& \left\|f(I^*)-f_{J}^{\mathbf{w^*}, \mathbf{b^*}}(I^*)\right\| \leq \left\|f(I^*)-f_{J}^{\mathbf{w}, \mathbf{b}}(I^*)\right\| \\
& + \left\|f_{J}^{\mathbf{w^*}, \mathbf{b^*}}(I^*)-f_{J}^{\mathbf{w}, \mathbf{b}}(I^*)\right\|
, \forall h=f_{J}^{\mathbf{w}, \mathbf{b}} \in \mathcal{H}^{J},
\end{aligned}
\end{equation}
where $\mathcal{H}^{J}$ represents the CNN space:
\begin{equation}
\mathcal{H}^{J}=\left\{h=f_{J}^{\mathbf{w}, \mathbf{b}} \mid \mathbf{w} \in \Phi_{\mathbf{w}}^J, \mathbf{b} \in \Phi_{\mathbf{b}}^J\right\},
\end{equation}
where the $\Phi_{\mathbf{w}}^J =\cup_{\tau=1}^J \cup_{\upsilon=1}^{s_\tau}\left\{w^{\tau \upsilon}\right\} $ and $\Phi_{\mathbf{b}}^J = \cup_{\tau=1}^J \cup_{\upsilon=1}^{s_\tau}\left\{b^{\tau \upsilon}\right\}$ represent weight space and bias space, respectively. And the $s_\tau$ represents the convolution kernel size of $\tau$-th convolution layer. Since we consider that the CNN has converged, we no longer consider the value of $\left\|f_{J}^{\mathbf{w^*}, \mathbf{b^*}}(I^*)-f_{J}^{\mathbf{w}, \mathbf{b}}(I^*)\right\|$, which means $f_{J}^{\mathbf{w}, \mathbf{b}}$ can be regarded as $f_{J}^{\mathbf{w^*}, \mathbf{b^*}}$ . Then the approximation accuracy only needs to consider $\left\|f(I^*)-f_{J}^{\mathbf{w}, \mathbf{b}}(I^*)\right\|$.

According to Formula (25), if $f_{J}^{\mathbf{w^*}, \mathbf{b^*}}$ approximates to $f$, it should have the following properties:
\begin{equation}
\| f_{J}^{\mathbf{w^*}, \mathbf{b^*}}(x_0) - f_{J}^{\mathbf{w}, \mathbf{b}}(x_1) \| \approx  \| f(x_0)-f(x_1)\| \geq \|x_0 - x_1\|.
\end{equation}

We use $Conv_1(\sigma(x; \Phi_{\mathbf{w}}^1 , \Phi_{\mathbf{b}}^1 ))$ to represent one convolution layer with ReLU function $\sigma$, then we can obtain:
\begin{equation}
\begin{aligned}
&\| Conv_1(\sigma(x_0; \Phi_{\mathbf{w^*}}^1 , \Phi_{\mathbf{b^*}}^1 )) - Conv_1(\sigma(x_1; \Phi_{\mathbf{w^*}}^1 , \Phi_{\mathbf{b^*}}^1 )) \| \\
&\geq \|x_0 - x_1\| .
\end{aligned}
\end{equation}

For a CNN with two convolution layers, we have:
\begin{equation}
\begin{aligned}
&\| Conv_2(\sigma(\textbf{O}_1^{x_0}; \Phi_{\mathbf{w^*}}^2 , \Phi_{\mathbf{b^*}}^2 )) - Conv_2(\sigma(\textbf{O}_1^{x_1};
\Phi_{\mathbf{w^*}}^2 , \Phi_{\mathbf{b^*}}^2 )) \| \\
&\geq \|\textbf{O}_1^{x_0} - \textbf{O}_1^{x_1}\| .
\end{aligned}
\end{equation}
where $\textbf{O}_1^{x_0}$ and $\textbf{O}_1^{x_1}$ represent the output of first layer convolution with $x_0$ and $x_1$ as the original input, respectively. Then, for $f_{J}^{\mathbf{w^*}, \mathbf{b^*}}$, the input of the $J$-th layer is the output of the $(J-1)$-th layer, we can obtain the general form:
\begin{equation}
\begin{aligned}
&\| Conv_J(\sigma(\textbf{O}_{J-1}^{x_0}; \Phi_{\mathbf{w^*}}^J , \Phi_{\mathbf{b^*}}^J )) - Conv_J(\sigma(\textbf{O}_{J-1}^{x_1};
\Phi_{\mathbf{w^*}}^J , \Phi_{\mathbf{b^*}}^J )) \| \\
&\geq \|\textbf{O}_{J-1}^{x_0} - \textbf{O}_{J-1}^{x_1}\| \geq \cdots \geq \|\textbf{O}_1^{x_0} - \textbf{O}_1^{x_1}\| \geq \|x_0 - x_1\| .
\end{aligned}
\end{equation}

If there exists $p \in { \{\|\textbf{O}_{J-1}^{x_0} - \textbf{O}_{J-1}^{x_1}\|, \dots,  \|\textbf{O}_1^{x_0} - \textbf{O}_1^{x_1}\| \} }$ satisfying $\| \|f(x_0)-f(x_1)\| - p \| \leq \mathcal{E} $, where $\mathcal{E}$ means a minimum, we will consider $f_{J}^{\mathbf{w^*}, \mathbf{b^*}}$ approximates to $f$. 

According to \cite{bouvrie2006notes} and \cite{rumelhart1986learning}, we solved the Formula (28) by updating $\mathbf{w}$ and $\mathbf{b}$ through backpropagation errors, and the scope of each update is limited. After a finite number of iterations for  Formula (28), the lower bound of $\| f_{J}^{\mathbf{w^*}, \mathbf{b^*}}(x_0) - f_{J}^{\mathbf{w^*}, \mathbf{b^*}}(x_1) \| $ is still determined by $\|x_0 - x_1\|$, which is as shown in  Formula (32). In our case, the positive gap is established by the border of glass. The pixel values within the boundry belong to $D_1$ and the pixel values outside the boundary belong to $D_0$. Then we replace $I^*$ by  $(FFT(I^*).real + I^*)$ as the input of $f_{J}^{\mathbf{w^*}, \mathbf{b^*}}$, the  Formula (30) can be reformed as:
\begin{equation}
\begin{aligned}
& \left\|f(I^*)-f_{J}^{\mathbf{w^*}, \mathbf{b^*}}(FFT(I^*).real + I^*)\right\| \\
& \leq \left\|f(I^*)-f_{J}^{\mathbf{w}, \mathbf{b}}(FFT(I^*).real + I^*))\right\| \\
& + \left\|f_{J}^{\mathbf{w^*}, \mathbf{b^*}}(FFT(I^*).real + I^*)-f_{J}^{\mathbf{w}, \mathbf{b}}(FFT(I^*).real + I^*)\right\| \\
& , \forall h=f_{J}^{\mathbf{w}, \mathbf{b}} \in \mathcal{H}^{J}.
\end{aligned}
\end{equation}

In the above, the FFT can amplify the differences of the pixel around the boundary in frequency domain, then we can obtain:
\begin{equation}
\begin{aligned}
&mean(\| x_0^{fft} - x_1^{fft} \|) > mean(\| x_0 - x_1 \|), \\
&(x_0^{fft}, x_1^{fft}) \in (FFT(I^*).real + I^*), \quad (x_0, x_1) \in I^*.
\end{aligned}
\end{equation}

After $(FFT(I^*).real + I^*)$ and $I^*$ are utilized as the inputs of the CNN, we used $p^{fft*}$ and $p^*$ to represent the maximum value in ${ \{\|\textbf{O}_{J-1}^{x_0} - \textbf{O}_{J-1}^{x_1}\|, \dots,  \|\textbf{O}_1^{x_0} - \textbf{O}_1^{x_1}\| \} }$, respectively. According to Formula (37), we can obtain:
\begin{equation}
\| \|f(x_0)-f(x_1)\| - p^{fft^*} \| < \| \|f(x_0)-f(x_1)\| - p^* \|
\end{equation}
which means:
\begin{equation}
	\left\|f(I^*)-f_{J}^{\mathbf{w}, \mathbf{b}}(FFT(I^*).real + I^*))\right\| < \left\|f(I^*)-f_{J}^{\mathbf{w}, \mathbf{b}}(I^*)\right\|.
\end{equation}

Then the Proposition 1 is proved.


According to the $\underset{\eta}{g_1}\left(x_{t r i}^{\text {ont }}\right)$ as weight branch $W$, the $\underset{\varpi}{g_2}(X_{En,3})$ as variable branch $\aleph$ and the $\underset{\xi}{g_3}\left(X_{De,1}\right)$ as bias branch $B$, we formally define learnable Fourier Convolution Controller as:
\begin{equation}
	FCC=\underset{\eta}{g_1}(x_{t r i}^{\text {out}}) \cdot [\underset{\varpi}{g_2}(X_{En,3})]^T+[\underset{\xi}{g_3}(X_{De},1)]^T.
\end{equation}

In the FCC, the $x_{t r i}^{\text {out }}$ leads and magnifies the significance of fine-grained focusing regions based on shallow and large objects in $X_{En,3}$ without over-capturing, besides $X_{De,1}$ utilizes the boundary constraint feature to flexibly correct and limit the position of semantic regions.

\begin{table*}[tbp]
	\centering
	\caption{Quantitative comparison to the state-of-the-art methods on the datasets of GDD \cite{9}, Trans10k-stuff \cite{1} and HSO \cite{10}. All the methods are re-trained on the corresponding training set. 
		$\bullet$: semantic segmentation methods. 
		$\circ$: salient object detection methods.
		$\triangle$: shadow detection methods.
		$\S$: medical image segmentation method.
		$\ast$: camouflaged object detection
		$\blacktriangle$: mirror segmentation methods. 
		$\diamond$: transparent object segmentation methods. 
		$\star$: glass segmentation methods. 
		The first and second best results are marked in red and blue, respectively. Our method achieves the best performance on all three challenging datasets under three standard metrics.
	}
\begin{tabular}{c|c|ccc|ccc|ccc}
		\hline
		\multirow{3}{*}{Methods} & \multirow{3}{*}{Pub.'Year} & \multicolumn{3}{c|}{GDD}                                                       & \multicolumn{3}{c|}{Trans10K-stuff}                                           & \multicolumn{3}{c}{HSO}                                                       \\ \cline{3-11} 
		&                            & \multicolumn{3}{c|}{Trainset:2980 Testset:936}                                 & \multicolumn{3}{c|}{Trainset:2455 Testset:1771}                                & \multicolumn{3}{c}{Trainset:3070 Testset:1782}                                \\ \cline{3-11} 
		&                            & \multicolumn{1}{c|}{IoU $\uparrow$}        & \multicolumn{1}{c|}{MAE$\downarrow$}   & BER$\downarrow$   & \multicolumn{1}{c|}{IoU$\uparrow$}        & \multicolumn{1}{c|}{MAE$\downarrow$}   & BER $\downarrow$  & \multicolumn{1}{c|}{IoU$\uparrow$}        & \multicolumn{1}{c|}{MAE$\downarrow$}   & BER $\downarrow$  \\ \hline
		ICNet$\bullet$                    & ECCV'18                    & \multicolumn{1}{c|}{69.78}  & \multicolumn{1}{c|}{0.163} & 15.93 & \multicolumn{1}{c|}{75.02}  & \multicolumn{1}{c|}{0.108} & 10.88 & \multicolumn{1}{c|}{62.38}  & \multicolumn{1}{c|}{0.163} & 17.01 \\
		PSPNet$\bullet$                   & CVPR'17                    & \multicolumn{1}{c|}{84.29} & \multicolumn{1}{c|}{0.082} & 8.73  & \multicolumn{1}{c|}{87.97}  & \multicolumn{1}{c|}{0.042} & 5.40  & \multicolumn{1}{c|}{77.74}  & \multicolumn{1}{c|}{0.092} & 10.48 \\
		DeepLabv3+$\bullet$               & ECCV'18                    & \multicolumn{1}{c|}{70.06} & \multicolumn{1}{c|}{0.142} & 15.21 & \multicolumn{1}{c|}{51.66}  & \multicolumn{1}{c|}{0.224} & 23.75 & \multicolumn{1}{c|}{64.69} & \multicolumn{1}{c|}{0.146} & 15.97 \\
		DenseASPP$\bullet$                & CVPR'18                    & \multicolumn{1}{c|}{83.75}  & \multicolumn{1}{c|}{0.078} & 8.61  & \multicolumn{1}{c|}{86.49}  & \multicolumn{1}{c|}{0.048} & 6.10  & \multicolumn{1}{c|}{76.01}  & \multicolumn{1}{c|}{0.092} & 11.28 \\
		BiSeNet$\bullet$                  & ECCV'18                    & \multicolumn{1}{c|}{80.31}  & \multicolumn{1}{c|}{0.102} & 10.99 & \multicolumn{1}{c|}{85.95}  & \multicolumn{1}{c|}{0.054} & 6.08  & \multicolumn{1}{c|}{75.97}  & \multicolumn{1}{c|}{0.100} & 11.02 \\
		DANet$\bullet$                    & CVPR'19                    & \multicolumn{1}{c|}{84.28}  & \multicolumn{1}{c|}{0.086} & 8.89  & \multicolumn{1}{c|}{88.26}  & \multicolumn{1}{c|}{0.043} & 5.21  & \multicolumn{1}{c|}{77.78}  & \multicolumn{1}{c|}{0.090} & 10.54 \\
		CCNet$\bullet$                    & ICCV'19                    & \multicolumn{1}{c|}{84.37}  & \multicolumn{1}{c|}{0.083} & 8.61  & \multicolumn{1}{c|}{88.29}  & \multicolumn{1}{c|}{0.042} & 5.10  & \multicolumn{1}{c|}{78.26}  & \multicolumn{1}{c|}{0.089} & 10.30 \\
		GFFNet$\bullet$                   & AAAI'20                    & \multicolumn{1}{c|}{82.41} & \multicolumn{1}{c|}{0.855}  & 9.11  & \multicolumn{1}{c|}{69.29}  & \multicolumn{1}{c|}{0.143} & 14.19 & \multicolumn{1}{c|}{77.34}  & \multicolumn{1}{c|}{0.094} & 9.69  \\
		SFNet$\bullet$                    & ECCV'20                    & \multicolumn{1}{c|}{81.04} & \multicolumn{1}{c|}{0.101} & 10.18 & \multicolumn{1}{c|}{71.34}  & \multicolumn{1}{c|}{0.130} & 13.08 & \multicolumn{1}{c|}{77.59}  & \multicolumn{1}{c|}{0.088} & 10.68 \\
		FaPN$\bullet$                     & ICCV'21                    & \multicolumn{1}{c|}{86.77}  & \multicolumn{1}{c|}{0.060} & 5.63  & \multicolumn{1}{c|}{89.12}  & \multicolumn{1}{c|}{\textcolor{blue}{0.041}} & 4.76  & \multicolumn{1}{c|}{78.16}  & \multicolumn{1}{c|}{0.086} & 9.48  \\ 
		PIDNet$\bullet$                      & CVPR’23                   & \multicolumn{1}{c|}{87.42}  & \multicolumn{1}{c|}{0.061} & 5.58  & \multicolumn{1}{c|}{77.68}  & \multicolumn{1}{c|}{0.094} & 9.27  & \multicolumn{1}{c|}{79.87}  & \multicolumn{1}{c|}{\textcolor{blue}{0.082}} & 9.47 \\ 
		BiRefNet$\bullet$                      & CAAI AIR' 24                   & \multicolumn{1}{c|}{89.15}  & \multicolumn{1}{c|}{0.052} & 5.05  & \multicolumn{1}{c|}{88.83}  & \multicolumn{1}{c|}{0.045} & 4.71  & \multicolumn{1}{c|}{79.98}  & \multicolumn{1}{c|}{0.083} & 9.43 \\ \hline
		DSS$\circ$                      & TPAMI'19                   & \multicolumn{1}{c|}{80.24}  & \multicolumn{1}{c|}{0.123} & 9.73  & \multicolumn{1}{c|}{84.77}  & \multicolumn{1}{c|}{0.075} & 6.42  & \multicolumn{1}{c|}{73.08}  & \multicolumn{1}{c|}{0.135} & 12.04 \\
		PiCANet$\circ$                  & CVPR'18                    & \multicolumn{1}{c|}{83.74}  & \multicolumn{1}{c|}{0.093} & 8.24  & \multicolumn{1}{c|}{83.99}  & \multicolumn{1}{c|}{0.077} & 7.03  & \multicolumn{1}{c|}{71.66} & \multicolumn{1}{c|}{0.148} & 13.31 \\
		RAS$\circ$                      & ECCV'18                    & \multicolumn{1}{c|}{81.05}  & \multicolumn{1}{c|}{0.104} & 9.43  & \multicolumn{1}{c|}{85.48}  & \multicolumn{1}{c|}{0.061} & 6.17  & \multicolumn{1}{c|}{74.76}  & \multicolumn{1}{c|}{0.113} & 11.18 \\
		CPD$\circ$                      & CVPR'19                    & \multicolumn{1}{c|}{82.62}  & \multicolumn{1}{c|}{0.092} & 8.81  & \multicolumn{1}{c|}{86.14}  & \multicolumn{1}{c|}{0.060} & 5.85  & \multicolumn{1}{c|}{76.24}  & \multicolumn{1}{c|}{0.108} & 10.49 \\
		EGNet$\circ$                    & ICCV'19                    & \multicolumn{1}{c|}{85.11} & \multicolumn{1}{c|}{0.081} & 7.39  & \multicolumn{1}{c|}{84.64}  & \multicolumn{1}{c|}{0.065} & 6.51  & \multicolumn{1}{c|}{74.37}  & \multicolumn{1}{c|}{0.116} & 11.55 \\
		F$^3$Net$\circ$                    & AAAI'20                    & \multicolumn{1}{c|}{84.87}  & \multicolumn{1}{c|}{0.078} & 7.33  & \multicolumn{1}{c|}{86.37}  & \multicolumn{1}{c|}{0.058} & 5.78  & \multicolumn{1}{c|}{76.95}  & \multicolumn{1}{c|}{0.102} & 10.53 \\
		MINet-R$\circ$                  & CVPR'20                    & \multicolumn{1}{c|}{82.17} & \multicolumn{1}{c|}{0.089} & 8.49  & \multicolumn{1}{c|}{85.94} & \multicolumn{1}{c|}{0.059} & 6.01  & \multicolumn{1}{c|}{76.65}  & \multicolumn{1}{c|}{0.104} & 10.27 \\
		ITSD$\circ$                     & CVPR'20                    & \multicolumn{1}{c|}{83.90}  & \multicolumn{1}{c|}{0.082} & 7.68  & \multicolumn{1}{c|}{85.62}  & \multicolumn{1}{c|}{0.062} & 6.23  & \multicolumn{1}{c|}{74.42} & \multicolumn{1}{c|}{0.120} & 11.36 \\
		LDF$\circ$                      & CVPR'20                   & \multicolumn{1}{c|}{83.36}    & \multicolumn{1}{c|}{0.085}      &     7.99  & \multicolumn{1}{c|}{84.53}            & \multicolumn{1}{c|}{0.067}      & 6.65      & \multicolumn{1}{c|}{76.97}     & \multicolumn{1}{c|}{0.102}      & 10.61      \\ 
		U2-Net$\circ$                      & PR'20                   & \multicolumn{1}{c|}{84.37}    & \multicolumn{1}{c|}{0.082}      &     8.54  & \multicolumn{1}{c|}{87.91}            & \multicolumn{1}{c|}{0.047}      & 5.08      & \multicolumn{1}{c|}{72.83}     & \multicolumn{1}{c|}{0.121}      & 12.41      \\ 
		Spider$\circ$                     & ICML'24               & \multicolumn{1}{c|}{86.19}      & \multicolumn{1}{c|}{0.080}  & 6.27    & \multicolumn{1}{c|}{87.36}           & \multicolumn{1}{c|}{0.047}      &  6.14  & \multicolumn{1}{c|}{74.70}          & \multicolumn{1}{c|}{0.136}      &     10.99  \\ \hline
		DSC$\triangle$                      & CVPR'19                    & \multicolumn{1}{c|}{83.56}  & \multicolumn{1}{c|}{0.090} & 7.97  & \multicolumn{1}{c|}{86.37}  & \multicolumn{1}{c|}{0.058} & 5.31  & \multicolumn{1}{c|}{71.93}  & \multicolumn{1}{c|}{0.128} & 13.11 \\
		DSD$\triangle$                      & CVPR'19                   & \multicolumn{1}{c|}{85.54}   & \multicolumn{1}{c|}{0.071}      &     7.17  & \multicolumn{1}{c|}{86.22}          & \multicolumn{1}{c|}{0.052}      &   5.57    & \multicolumn{1}{c|}{76.49}    & \multicolumn{1}{c|}{0.101}      &    11.07   \\
		BDRAR$\triangle$                    & ECCV'18                    & \multicolumn{1}{c|}{80.01}  & \multicolumn{1}{c|}{0.099} & 9.87  & \multicolumn{1}{c|}{85.00}  & \multicolumn{1}{c|}{0.061} & 6.04  & \multicolumn{1}{c|}{75.32}  & \multicolumn{1}{c|}{0.103} & 11.13 \\ \hline
		PraNet$\S$                   & MICCAI'20                  & \multicolumn{1}{c|}{82.06}  & \multicolumn{1}{c|}{0.098} & 9.33  & \multicolumn{1}{c|}{87.15}  & \multicolumn{1}{c|}{0.058} & 5.31  & \multicolumn{1}{c|}{71.93}  & \multicolumn{1}{c|}{0.128} & 13.11 \\ 
		U-Net$\S$                     & MICCAI'15               & \multicolumn{1}{c|}{75.44}      & \multicolumn{1}{c|}{0.136}  & 11.16    & \multicolumn{1}{c|}{70.95}           & \multicolumn{1}{c|}{0.135}      &  13.82  & \multicolumn{1}{c|}{60.23}          & \multicolumn{1}{c|}{0.201}      &     20.88  \\ 
		Transfuse$\S$                     & MICCAI'21               & \multicolumn{1}{c|}{81.78}      & \multicolumn{1}{c|}{0.092}  & 9.68    & \multicolumn{1}{c|}{88.6}           & \multicolumn{1}{c|}{0.042}      &  4.81  & \multicolumn{1}{c|}{77.6}          & \multicolumn{1}{c|}{0.108}      &     10.07  \\ \hline
		SINet$\ast$                    & CVPR'20               & \multicolumn{1}{c|}{79.28}       & \multicolumn{1}{c|}{0.133} & 10.14 & \multicolumn{1}{c|}{85.01}           & \multicolumn{1}{c|}{0.064}      & 6.37  & \multicolumn{1}{c|}{76.07}          & \multicolumn{1}{c|}{0.113}      &    10.92   \\
		PFNet$\ast$                     & CVPR'21               & \multicolumn{1}{c|}{85.96}      & \multicolumn{1}{c|}{0.072}  & 6.50    & \multicolumn{1}{c|}{85.68}           & \multicolumn{1}{c|}{0.059}      &  5.97  & \multicolumn{1}{c|}{76.79}          & \multicolumn{1}{c|}{0.109}      &     10.38  \\
		RankNet$\ast$                   & CVPR'21               & \multicolumn{1}{c|}{84.54}          & \multicolumn{1}{c|}{0.085}   &  7.70     & \multicolumn{1}{c|}{86.55}            & \multicolumn{1}{c|}{0.055}      & 5.77   & \multicolumn{1}{c|}{76.88}            & \multicolumn{1}{c|}{0.105}      &   10.59    \\ \hline
		MirrorNet$\blacktriangle $                & ICCV'19                    & \multicolumn{1}{c|}{85.07}  & \multicolumn{1}{c|}{0.083} & 7.67  & \multicolumn{1}{c|}{88.30}  & \multicolumn{1}{c|}{0.047} & 4.95  & \multicolumn{1}{c|}{78.82}  & \multicolumn{1}{c|}{0.102} & 9.93  \\
		PMD$\blacktriangle $                      & CVPR'20                    & \multicolumn{1}{c|}{87.12}        & \multicolumn{1}{c|}{0.066}      &     5.97  & \multicolumn{1}{c|}{88.10}          & \multicolumn{1}{c|}{0.047}      &  5.02     & \multicolumn{1}{c|}{\textcolor{blue}{80.31}}         & \multicolumn{1}{c|}{0.089}      &   \textcolor{blue}{8.81}    \\ \hline
		TransLab$\diamond $                 & ECCV'20                    & \multicolumn{1}{c|}{81.67}  & \multicolumn{1}{c|}{0.096} & 9.69  & \multicolumn{1}{c|}{87.11}  & \multicolumn{1}{c|}{0.051} & 5.44  & \multicolumn{1}{c|}{74.32}  & \multicolumn{1}{c|}{0.123} & 12.00 \\
		Trans2Seg$\diamond $                & IJCAI'21                   & \multicolumn{1}{c|}{84.65} & \multicolumn{1}{c|}{0.074} & 7.32  & \multicolumn{1}{c|}{75.01}  & \multicolumn{1}{c|}{0.122} & 10.66 & \multicolumn{1}{c|}{78.06}  & \multicolumn{1}{c|}{0.092} & 9.59  \\ \hline
		GDNet$\star $                    & CVPR'20                    & \multicolumn{1}{c|}{87.64}  & \multicolumn{1}{c|}{0.063} & 5.61  & \multicolumn{1}{c|}{88.72}  & \multicolumn{1}{c|}{0.045} & 4.71  & \multicolumn{1}{c|}{78.84}  & \multicolumn{1}{c|}{0.092} & 9.28  \\
		GSD$\star $                      & CVPR'21                    & \multicolumn{1}{c|}{87.62}  & \multicolumn{1}{c|}{0.065} & 5.88  & \multicolumn{1}{c|}{89.71}  & \multicolumn{1}{c|}{\textcolor{blue}{0.041}} & 4.49  & \multicolumn{1}{c|}{78.95}  & \multicolumn{1}{c|}{0.099} & 9.70  \\
		EBLNet$\star $                   & ICCV'21                    & \multicolumn{1}{c|}{84.85}            & \multicolumn{1}{c|}{0.079}      &   7.60    & \multicolumn{1}{c|}{\textcolor{red}{89.92}}            & \multicolumn{1}{c|}{0.047}      &    \textcolor{red}{4.29}   & \multicolumn{1}{c|}{79.21}        & \multicolumn{1}{c|}{0.094}      & 9.48      \\
		PGSNet$\star $                   & IEEE TIP'22                & \multicolumn{1}{c|}{87.81}  & \multicolumn{1}{c|}{0.062} & {5.56}  & \multicolumn{1}{c|}{\textcolor{blue}{89.79}}  & \multicolumn{1}{c|}{0.042} & \textcolor{blue}{4.39}  & \multicolumn{1}{c|}{80.06}  & \multicolumn{1}{c|}{0.089} & 9.08 \\
		RFENet$\star $                    & IJCAI'23                    & \multicolumn{1}{c|}{77.74}  & \multicolumn{1}{c|}{0.118} & 10.90  & \multicolumn{1}{c|}{81.60}  & \multicolumn{1}{c|}{0.082} & 8.17  & \multicolumn{1}{c|}{65.43}  & \multicolumn{1}{c|}{0.152} & 15.09  \\ 
		GateNet$\star $		& IJCV'24			& \multicolumn{1}{c|}{\textcolor{blue}{89.80}}  & \multicolumn{1}{c|}{\textcolor{blue}{0.049}} & \textcolor{blue}{4.90}  & \multicolumn{1}{c|}{-}  & \multicolumn{1}{c|}{-} & -  & \multicolumn{1}{c|}{-}  & \multicolumn{1}{c|}{-} & -  \\\hline
		FBWC              & Ours                        & \multicolumn{1}{c|}{\textcolor{red}{93.17}}          & \multicolumn{1}{c|}{\textcolor{red}{0.035}}      &  \textcolor{red}{4.03}     & \multicolumn{1}{c|}{89.74}            & \multicolumn{1}{c|}{\textcolor{red}{0.040}}      &  {4.51}     & \multicolumn{1}{c|}{\textcolor{red}{87.59}}            & \multicolumn{1}{c|}{\textcolor{red}{0.059}}      &    \textcolor{red}{6.27}   \\ \hline
	\end{tabular}
	\label{tab1}
\end{table*}

\section{Experiments}
\subsection{Experiment Setting}
\textbf{\textit{Datasets.}} For the performance evaluation of the proposed method, we carry out detailed experiments on three glass segmentation tasks: GDD \cite{9}, HSO \cite{10}, and Trans10K-Stuff \cite{1}. To provide further clarity, the GDD comprises a total of 3916 images. Among these, 2980 images are allocated to the train set, while the remaining 936 images constitute the test set. Simultaneously, HSO serves as a comprehensive scene segmentation dataset, primarily focusing on family environments. It is meticulously crafted to unlock a myriad of applications for domestic robots. The dataset divided 3070 of 4852 images into a training set and 1782 as a test set. Aiming at exploring the glass segmentation task, in the selection of the data set, we prefer to choose the segmented transparent glass targets. Thus, on the large-scale transparent object segmentation dataset Trans10K-Stuff, which contains two categories: stuff and things, we utilize the category of stuff in the whole dataset. The filtered result dataset Trans10K-Stuff consists of 2455 train set and 1771 test set. For training, input images are augmented by randomly horizontal flipping and resizing. Throughout the testing process, all datasets of images are resized into $352 \times 352$ without any pre-processing.

\textbf{\textit{Evaluation Metrics.}} The evaluation metrics used in the experiments for comparison of our proposed methods with other SOTA methods including, intersection over union (IoU), mean absolute error (MAE), and balanced error rate (BER), which are defined as follows:
\begin{equation}
	IoU=\frac{TP}{TP+FP+FN},
\end{equation}
\begin{equation}
	MAE=\frac{1}{H \times W} \sum_{(x, y)}^{(H, W)}|o(x, y)-l(x, y)|,
\end{equation}
\begin{equation}
	BER=100 \times\left(1-\frac{1}{2}\left(\frac{T P}{N_p}+\frac{T N}{N_n}\right)\right),
\end{equation}
where the value of true-positive is denoted as TP, false-positive as FP, true-negative as TN, and false-negative as FN respectively. Simultaneously, $o(x, y)$ is denoted as the output of models for glass prediction and $l(x, y)$ represents the labels of annotation on the pixel $(x, y)$ . $N_p$ and $N_n$ represent pixels in the images that contain and do not contain the segmented objects of glass, respectively.

\textbf{\textit{Implementation Details.}} The method outlined in our paper is implemented using the PyTorch framework on an NVIDIA RTX 4090 GPU equipped with 24GB memory. We run for 200 epochs with a batch size of 6 for the GPU. Before fine-tuning our models, we utilize pre-training weight as most of previous works, which is trained by ImageNet\cite{russakovsky2015imagenet}. The three public glass segmentation datasets are optimized using an SGD optimizer with a weight decay of $5 \times 10^{-4}$, an initial learning rate of 0.001, and a momentum of 0.9. Throughout the training process, we utilize the polynomial learning rate policy to gradually adjust the initial learning rate, denoted as $(1-\frac{{iter}}{{ max_{iter} }})^{0.9}$.

\subsection{Comparison Results}
In this study, to verify the effectiveness and robustness of the proposed method for the task of glass segmentation, the method is compared with a variety of methods in several related fields with few glass segmentation methods at present. We conduct extensive experiments to compare the segmentation performance of our work and state-of-the-art methods including PraNet \cite{33}, U-Net \cite{34}, and Transfuse \cite{35} on medical image segmentation methods. DSS \cite{36}, PiCANet \cite{37}, RAS \cite{38}, CPD \cite{39}, EGNet \cite{40}, F3Net \cite{41}, MINet-R \cite{42}, ITSD \cite{43}, LDF \cite{44}, U2-Net \cite{45} and Spider \cite{zhao2024spider} on salient object detection methods. ICNet \cite{46}, PSPNet \cite{47}, DeepLabv3+ \cite{48}, DenseASPP \cite{49}, BiSeNet \cite{50}, DANet \cite{52}, CCNet \cite{53}, GFFNet \cite{54}, SFNet \cite{55}, FaPN \cite{56}, PIDNet \cite{57} and BiRefNet \cite{zheng2024bilateral} on semantic segmentation methods. DSC \cite{58}, DSD \cite{59}, and BDRAR \cite{60} on shadow detection methods. SINet \cite{61}, PFNet \cite{62}, and RankNet \cite{63} on camouflaged object detection methods. MirrorNet \cite{64} and PMD \cite{8} on mirror segmentation methods. TransLab \cite{1} and Trans2Seg \cite{67} on transparent object segmentation methods. GDNet \cite{9}, GSD \cite{5}, EBLNet \cite{2}, PGSNet \cite{10}, RFENet \cite{RFENet} and GateNet\cite{zhao2024towards} on glass segmentation methods. To be fair, we retrained either their open codes or the implementations with recommended parameter settings on the three datasets for effective comparison.

\begin{figure*}
	\begin{center}
		\includegraphics[width = 1\linewidth]{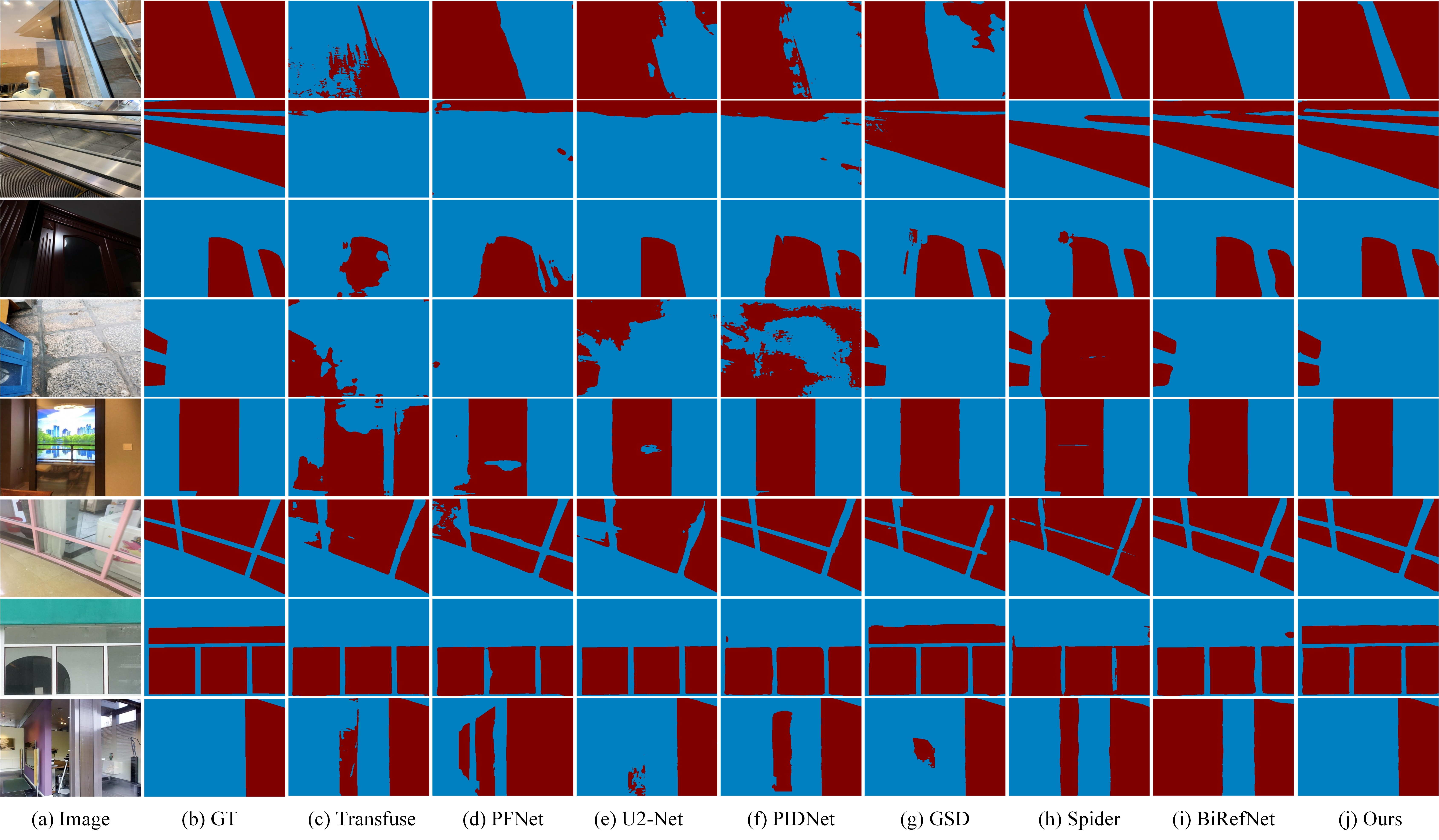}
	\end{center}
	\vspace{-18pt}
	\caption{Visual comparison of our FBWC with state-of-the-art methods.}
	\label{fig7}
	\vspace{-12pt}
\end{figure*}

Our quantitative results on the glass segmentation task achieve excellent performance compared to other 39 previous works on three benchmark datasets, which is present in Table \ref{tab1}. The proposed method achieves the best comprehensive performance and surpasses the SOTA method PGSNet \cite{10} on GDD with a significant improvement of IoU by 5.36\%, a decrement of MAE and BER by 0.025 and 1.53 respectively. Besides, the proposed method shows a slight advantage on Trans10K-stuff with MAE and can be consistent with the SOTA methods on other metrics. At the same time, we can also see that FBWC outperforms the previous works on all metrics based on HSO. Specifically, FBWC achieves 87.59\% on IoU, 0.059 on MAE, and 6.27 on BER, which are 7.28\% higher, 0.023 and 2.54 lower than the performance of the best glass segmentation method. From the test indexes of the three data sets, our FBWC comprehensively embodies excellent segmentation performance and generalization ability.

In addition to quantitative comparisons, from the visual qualitative results in Figure \ref{fig7}, we can observe that our FBWC can first fully constrain the boundary information of the segmented target region so that the visual results embrace a clear and complete segmentation boundary to distinguish the background and the glass region. Besides, our work relies on the advantages of shallow space acquisition structures to efficiently avoid the impact of over-capturing as noise on segmentation results. Finally, by establishing a semantic association between fine-grained feature focusing and large-area target region, the proposed method can accurately anchor the segmentation region to eliminate interference information.

\begin{table}[]
	\caption{Quantitative ablation results among different numbers of CUs. The Base-X means FBWC without WCC.}
	\label{tab2}
	\centering
	\begin{tabular}{c|cl|ccc}
		\hline 
		\multirow{2}{*}{Datasets} & \multicolumn{2}{c|}{\multirow{2}{*}{Methods}} & \multicolumn{3}{c}{Results} \\ \cline{4-6} 
		& & & IoU$\uparrow$   & MAE$\downarrow$  & BER$\downarrow$  \\ \hline
		& A                     & Base-X $+$ 1 CU                    & 78.98 & 0.099 & 9.85 \\
		&	B                     & Base-X $+$ 2 CUs         & 80.76 & 0.088 & 9.29 \\
		HSO &	C                     & Base-X $+$ 3 CUs      & 81.27 & 0.080 & 8.97 \\
		&	D                     & Base-X $+$ 4 CUs              & \textcolor{red}{87.59} & \textcolor{red}{0.059} & \textcolor{red}{6.27} \\ 
		&	E                     & Base-X $+$ 5 CUs          & 80.22 & 0.092 & 9.35  \\ \hline
		&		A                     & Base-X $+$ 1 CU                       & 78.54 & 0.082 & 8.87 \\
		&	B                     & Base-X $+$ 2 CUs          & 80.33 & 0.079 & 7.85 \\
		Trans10K-stuff &	C                     & Base-X $+$ 3 CUs        & 85.71 & 0.061 & 6.24 \\
		&	D                     & Base-X $+$ 4 CUs                & \textcolor{red}{89.74} & \textcolor{red}{0.040} & \textcolor{red}{4.51} \\ 
		&	E                     & Base-X $+$ 5 CUs          & 82.39 & 0.072 & 6.86  \\ \hline
		&		A                     & Base-X $+$ 1 CU                        & 88.38 & 0.062 & 5.58 \\
		&	B                     & Base-X $+$ 2 CUs          & 89.01 & 0.061 & 5.47 \\
		GDD &	C                     & Base-X $+$ 3 CUs       & 90.52 & 0.058 & 5.38 \\
		&	D                     & Base-X $+$ 4 CUs                & \textcolor{red}{93.17} & \textcolor{red}{0.035} & \textcolor{red}{4.03} \\ 
		&	E                     & Base-X $+$ 5 CUs         & 87.89 & 0.071 & 5.70 \\ \hline 
	\end{tabular}
\end{table}

\begin{figure}[tbp]
	\begin{center}
		\includegraphics[width=1\linewidth]{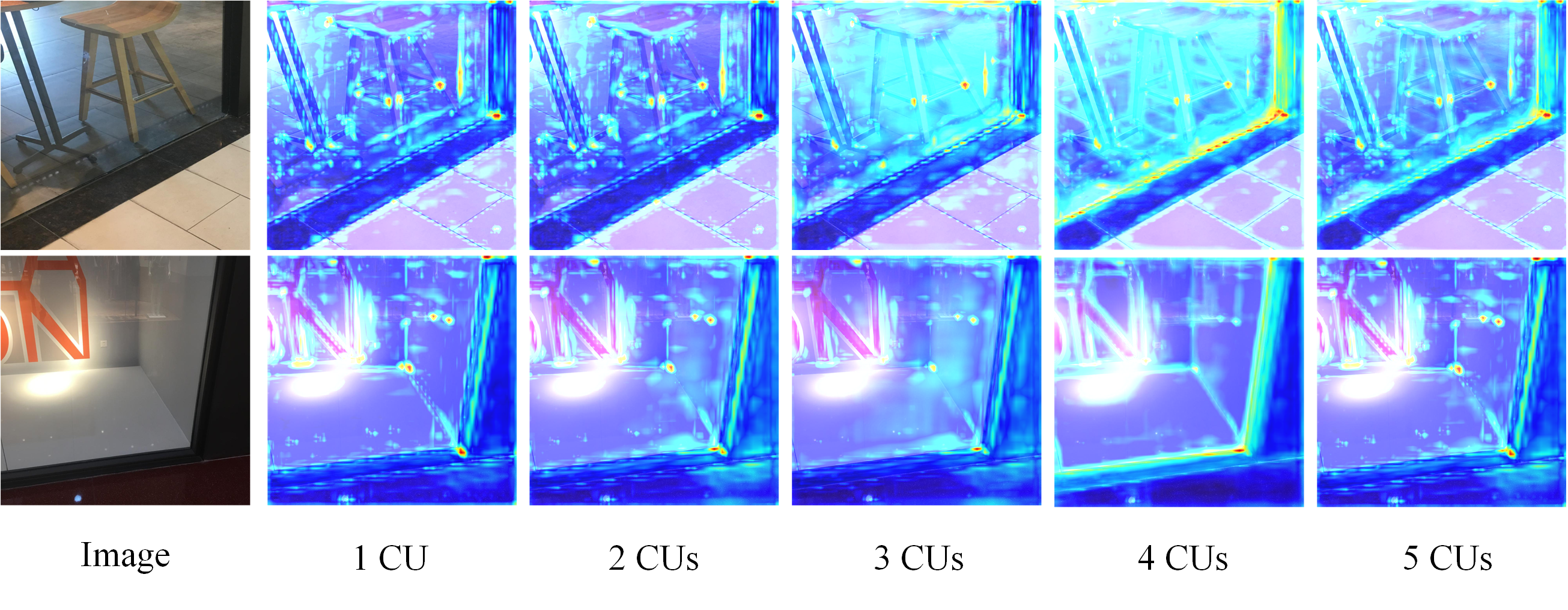}
	\end{center}
	\vspace{-18pt}
	\caption{Heat maps from different numbers of CUs.}
	\label{fig8}
	\vspace{-12pt}
\end{figure}

\subsection{Ablation Study}
To evaluate the ability of horizontal shallow branches in the glass segmentation task to avoid excessive extraction and the influence of various components of our proposed FBWC, we further conduct ablation studies on three datasets, which can verify the effectiveness and generalization ability of designed factors in the model.
Firstly, we design WCC against the phenomenon of over-capturing. With the increase of the depth of the network structure, semantic information will be gradually abstracted, and excessive detail noise will be prone to be introduced in the glass segmentation situation, which is not conducive to the complete acquisition of large-area targets. Due to the disadvantages of vertical structure, we concatenate CUs horizontally to avoid excessive abstraction of glass information. At the same time, the amount of CUs in the transverse connection process will also affect the experimental results. Therefore, we conducted ablation experiments on the numbers of CUs to determine the optimal quantity.

\subsubsection{\textbf{Numbers of CUs}}
Based on Table \ref{tab2}, we can observe a positive increase in all three evaluation metrics with the increment of CU numbers. This trend underscores the significance of the WCC within our FBWC, which comprises four distinct CUs. Within each CU, boundary constraints are introduced to ensure clear partition edges. Therefore, in the process of the gradual increase of the number of CUs, the continuous strengthening of constraints and the effective transfer of accumulated shallow information lead to improved segmentation results. However, this horizontal information transfer will lead to the last CU retaining miscellaneous information without providing more effective semantic information, and lead to lower segmentation accuracy. 

It can be seen from the heat map that with the increase in the number of CUs, the boundary constraints continue to strengthen, which is expressed in the Figure \ref{fig8} as the segmentation boundary gradually becomes obvious. When the number is from 4 to 5, the boundary constraints will degenerate due to the interference of redundant information. Based on the quantitative results from the numbers 4 to 5, we prudently chose 4 as the optimal number to balance the effects of boundary constraint validity and semantic information redundancy on segmentation results.

\subsubsection{\textbf{Trough Point}}
Considering the vertical structure, in the CU, we design a trough point on the third layer for properly capturing spatial information without further deepening. To verify the appropriateness of the occurrence of trough points, trough points on different layers are designed into the ablation experiment.

\begin{table}[t]
	\caption{Quantitative ablation results among different layers of trough point.}
	\label{tab3}
	\centering
	\begin{tabular}{c|cl|ccc}
		\hline 
		\multirow{2}{*}{Datasets} & \multicolumn{2}{c|}{\multirow{2}{*}{Methods}} & \multicolumn{3}{c}{Results} \\ \cline{4-6} 
		& & & IoU$\uparrow$   & MAE$\downarrow$  & BER$\downarrow$  \\ \hline
	
		&	A                     & the second layer        & 78.64 & 0.101 & 10.07 \\
		
		HSO &	B                     & the third layer             & \textcolor{red}{87.59} & \textcolor{red}{0.059} & \textcolor{red}{6.27} \\ 
		&	C                     & the fourth layer         & 81.52 & 0.078 & 8.81  \\ \hline
		&	A                     & the second layer          & 75.26 & 0.106 & 9.34 \\
		
		Trans10K-stuff &	B                     & the third layer              & \textcolor{red}{89.74} & \textcolor{red}{0.040} & \textcolor{red}{4.51} \\ 
		&	C                     & the fourth layer          & 80.76 & 0.077 & 7.79  \\ \hline
		&	A                     & the second layer          & 87.22 & 0.067 & 5.71 \\
		
		GDD &	B                     & the third layer               & \textcolor{red}{93.17} & \textcolor{red}{0.035} & \textcolor{red}{4.03} \\ 
		&	C                     & the fourth layer          & 87.02 & 0.068 & 5.78  \\ \hline 

	\end{tabular}
\end{table}

As shown in Table \ref{tab3}, trough points on the second layer under the three datasets perform worse than the other experimental groups based on a small number of convolutions. Therefore, the advanced appearance of the trough point will cause WCC to over-capture shallow semantic information and fail to excavate the intra-class consistency features without further abstraction. By contrast, when extending and deepening trough points into the fourth layer, the segmentation performance of the model is negatively affected by multi-layer convolution to over-abstracting dispersive fine-grained features as noise. In general, faced with the glass segmentation task, an appropriate trough point can capture the commonalities among glass classes through the appropriate convolution quantity while avoiding the difficulty of feature clustering and the decrease of semantic relevance caused by excessive extraction. Similarly, the proposed CTA and FCC are also considered in the ablation experiment to verify the validity of the designed attention mechanism and Fourier enhancement for glass features.

\begin{figure}[t]
	\begin{center}
		\includegraphics[width=1\linewidth]{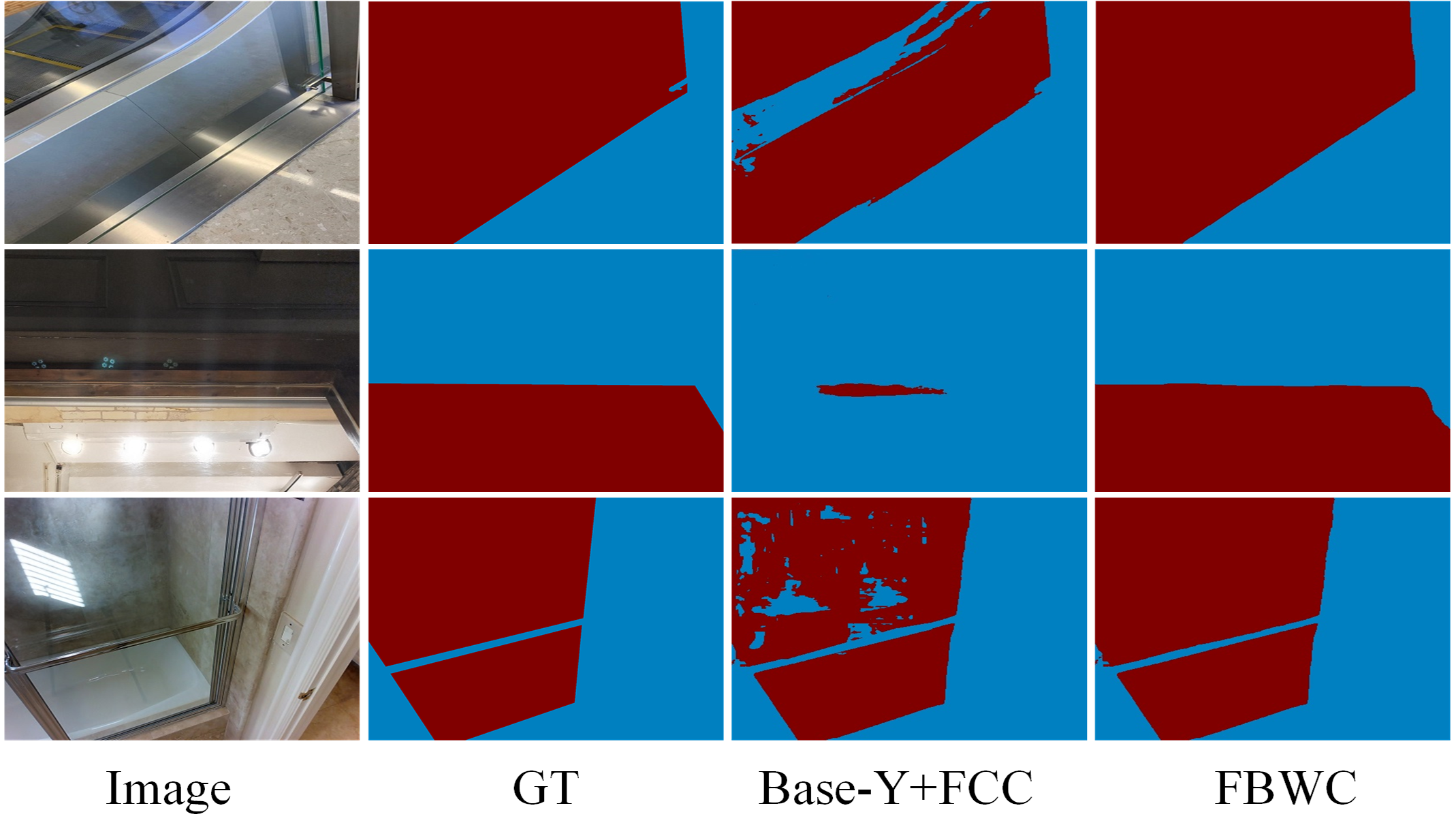}
	\end{center}
	\vspace{-12pt}
	\caption{Visual comparison of ablation results about CTA. Base-Y+FCC: FBWC without CTA.}
	\label{fig9}
\end{figure}
\begin{table}[tp]
	\caption{Quantitative ablation results of BC Loss and AM Loss. }
	\label{tab10}
	\centering
	\begin{tabular}{c|cc|ccc}
		\hline 
		\multirow{2}{*}{Datasets} & \multicolumn{2}{c|}{\multirow{2}{*}{Methods}} & \multicolumn{3}{c}{Results} \\ \cline{4-6} 
		& & & IoU$\uparrow$   & MAE$\downarrow$  & BER$\downarrow$  \\ \hline

		&	A                     &  BC Loss (w/o)         &83.60 & 0.081 & 8.24 \\
		HSO&	B          &  AM Loss (w/o)    & 87.15 & 0.064 & 6.40  \\
		&	C          &  FBWC   & \textcolor{red}{87.59} & \textcolor{red}{0.059} & \textcolor{red}{6.27}  \\ \hline
		
		&	A       &   BC Loss (w/o)        & 83.08 & 0.068 & 6.75 \\
		Trans10K-stuff&	B         & AM Loss (w/o)        & 87.97 & 0.051 & 4.97\\
		&	C         & FBWC      & \textcolor{red}{89.74} & \textcolor{red}{0.040} & \textcolor{red}{4.51} \\ \hline

		&	A            &   BC Loss (w/o)      & 88.54 & 0.061 & 6.32 \\
		GDD&	B           & AM Loss (w/o)           & 90.94 & 0.048 & 5.13 \\ 
		&	C          & FBWC         & \textcolor{red}{93.17} & \textcolor{red}{0.035} & \textcolor{red}{4.03} \\ \hline
	
	\end{tabular}
\end{table}

\begin{table}[tp]
	\caption{Quantitative ablation results of CA, CTA, SCC and FCC. The Base-Y means FBWC without CA, CTA, SCC and FCC.}
	\label{tab4}
	\centering
	\begin{tabular}{c|cc|ccc}
		\hline 
		\multirow{2}{*}{Datasets} & \multicolumn{2}{c|}{\multirow{2}{*}{Methods}} & \multicolumn{3}{c}{Results} \\ \cline{4-6} 
		& & & IoU$\uparrow$   & MAE$\downarrow$  & BER$\downarrow$  \\ \hline

		&	A                     & Base-Y $+$ SCC        & 79.93 & 0.092 & 9.05 \\
		&	B                    & Base-Y $+$ FCC        & 81.66 & 0.079 & 8.76 \\
		HSO &	C                    & Base-Y $+$   CA           & 83.78 & 0.086 & 7.73 \\ 
		&      D    & Base-Y $+$ CTA            & 84.07 & 0.079 & 7.61 \\ 
		&	E                    & FBWC         & \textcolor{red}{87.59} & \textcolor{red}{0.059} & \textcolor{red}{6.27}  \\ \hline

		&	A                     & Base-Y $+$ SCC           & 81.57 & 0.083 & 6.98 \\
		&	B                     & Base-Y $+$ FCC           & 83.48 & 0.068 & 6.73 \\
		Trans10K-stuff &	C                  & Base-Y $+$   CA           & 85.33 & 0.062 & 5.68 \\     
		&      D   & Base-Y $+$ CTA            & 86.91 & 0.058 & 5.54 \\ 
		&	E                    & FBWC           & \textcolor{red}{89.74} & \textcolor{red}{0.040} & \textcolor{red}{4.51}  \\ \hline

		&	A                     & Base-Y $+$ SCC          & 88.64 & 0.065 & 5.45 \\
		&	B                     & Base-Y $+$ FCC          & 90.85 & 0.058 & 5.29 \\
		GDD &	C             & Base-Y $+$   CA           & 91.98 & 0.043 & 4.52 \\          
		&      D   & Base-Y $+$ CTA              & 92.79 & 0.037 & 4.29 \\ 
		&	E                     & FBWC           & \textcolor{red}{93.17} & \textcolor{red}{0.035} & \textcolor{red}{4.03}  \\ \hline
		
	\end{tabular}
\end{table}

\begin{table}[b]
	\caption{Quantitative ablation results of FBWC without CTA on different number of CUs. The Base-Z means FBWC without WCC, CTA and FCC.}
	\label{tab5}
	\centering
	\begin{tabular}{c|cl|cc}
		\hline 
		\multirow{2}{*}{Datasets} & \multicolumn{2}{c|}{\multirow{2}{*}{Methods}} & \multicolumn{2}{c}{Results} \\ \cline{4-5} 
		& & & IoU$\uparrow$   & MAE$\downarrow$  \\ \hline
		
		&	A                     & Base-Z $+$ FCC $+$ 1 CU           & 72.77 & 0.126  \\
		&	B                     & Base-Z $+$ FCC $+$ 2 CUs           & 74.38 & 0.123  \\
		HSO &	C                     & Base-Z $+$ FCC $+$ 3 CUs              & 75.46 & 0.115 \\ 
		&	D                     & Base-Z $+$ FCC $+$ 4 CUs            & \textcolor{red}{81.66} & \textcolor{red}{0.079} \\ 
		&	E                     & Base-Z $+$ FCC $+$ 5 CUs           & 76.02 & 0.103  \\ \hline 
		
		&	A                     & Base-Z $+$ FCC $+$ 1 CU          & 75.33 & 0.092  \\
		&	B                     & Base-Z $+$ FCC $+$ 2 CUs          & 77.49 & 0.086  \\
		
		Trans10K-stuff &	C                     & Base-Z $+$ FCC $+$ 3 CUs             & 79.53 & 0.080  \\ 
		&	D                     & Base-Z $+$ FCC $+$ 4 CUs            & \textcolor{red}{83.48} & \textcolor{red}{0.068} \\ 
		&	E                     & Base-Z $+$ FCC $+$ 5 CUs            & 78.05 & 0.084  \\ \hline
		
		&	A                     & Base-Z $+$ FCC $+$ 1 CU        & 86.72 & 0.071  \\
		&	B                     & Base-Z $+$ FCC  $+$ 2 CUs       & 87.24 & 0.068  \\
		
		GDD &	C                     & Base-Z $+$ FCC $+$ 3 CUs             & 88.71 & 0.062  \\ 
		&	D                     & Base-Z $+$ FCC $+$ 4 CUs         & \textcolor{red}{90.85} & \textcolor{red}{0.058}  \\ 
		&	E                     & Base-Z $+$ FCC $+$ 5 CUs        & 86.39 & 0.073  \\ \hline 
		
	\end{tabular}
\end{table}

\subsubsection{\textbf{Designed Module and Loss}} 
From Table \ref{tab10}, we can observe the supervisory effects of BC Loss and AM Loss on the model. Specifically, when FBWC has no boundary constraints, the model's ability to learn boundary features of the segmentation target weakens, resulting in a significant decrease in model performance. In addition, AM Loss can effectively improve the semantic feature consistency of large-scale segmentation regions in the model, thereby enhancing the coherence and naturalness of the segmentation results.

Based on Table \ref{tab4}, we observe that the segmentation performance on all three datasets decreased without CTA to supplement the focused semantic features and maintain the integrity of semantic information in the edge constraint region. From Figure \ref{fig9}, it can be vividly seen that CTA can effectively filter the reflected strong light noise, fully focus the semantic information in the target region and reduce the dispersion of segmentation results. In order to verify the effectiveness of transposition operation in CTA, we removed the transposition process in the Cross Transpose Attention (CTA) module and made it a Cross Attention (CA) module. From Table \ref{tab4}, it can be seen that the comprehensive performance of the CTA module on the three datasets is better than that of CA, indicating that transposition operation can establish effective global correlations among glass semantic features and boost segmentation performance.

\begin{figure}[t]
	\begin{center}
		\includegraphics[width=1\linewidth]{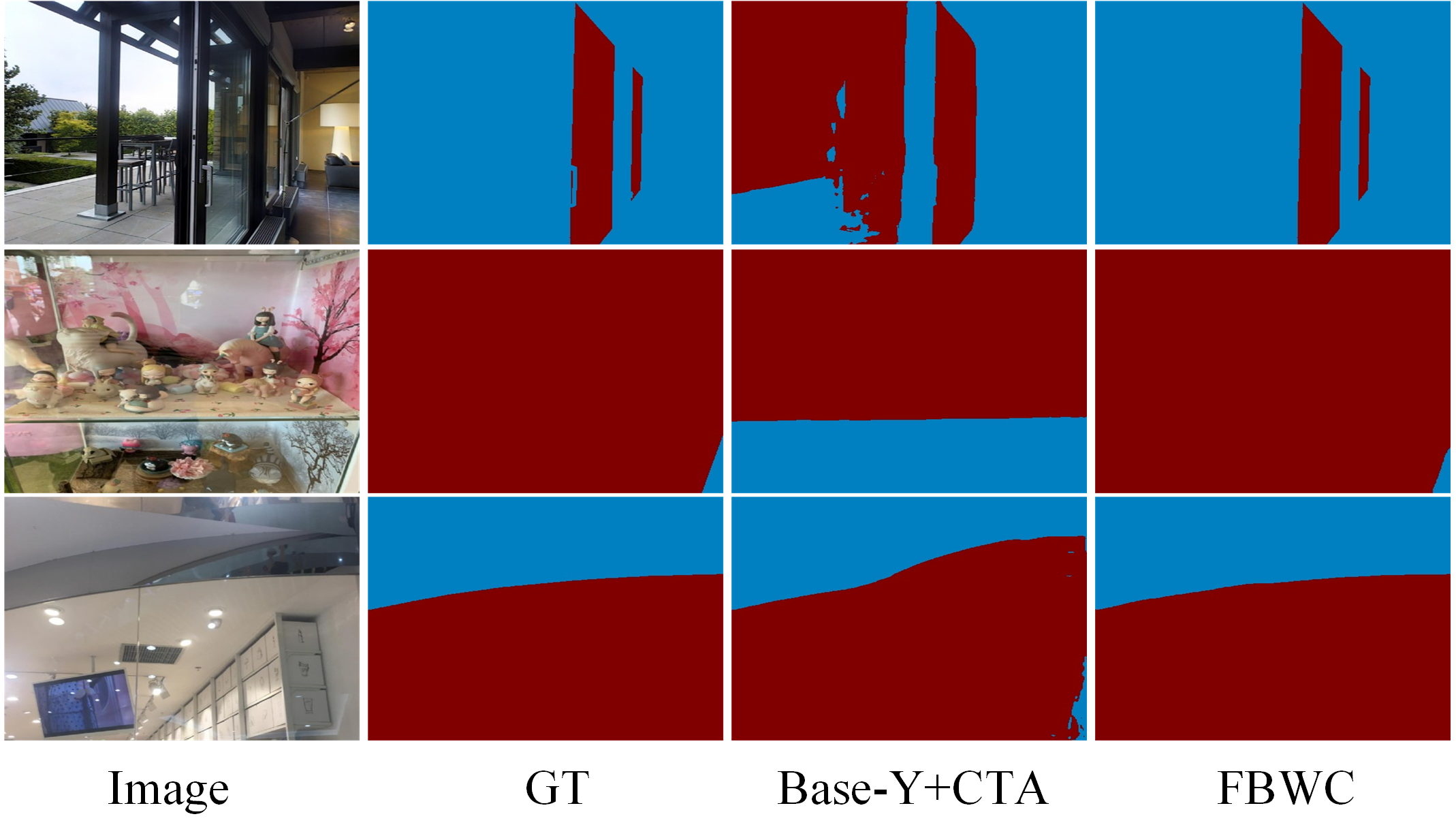}
	\end{center}
	\vspace{-12pt}
	\caption{Visual comparison of ablation results about FCC. Base-Y+CTA : FBWC without FCC.}
	\label{fig10}
\end{figure}

At the same time, we introduce the Fourier Transform on a variable branch to FCC, which is conducive to further strengthening the boundary information, increasing the boundary sensitivity of the model, and improving the segmentation accuracy. Visual results in Figure \ref{fig10} show the importance of the FCC, which observes a sharp gradient decline of edge pixels in the conversion process of the frequency domain and spatial domain. Finally, the boundary information is integrated and strengthened to effectively divide the boundary between glass and reality. Moreover, We designed relevant ablation experiments for this, specifically replacing FFT with a standard convolution operation to obtain Standard Convolution Controller (SCC). From Table V, it can be seen that SCC performs worse than FCC in the datasets HSO, Trans10K-stuff and GDD.

\begin{table}[t]
	\caption{Quantitative ablation results of FBWC without FCC on different number of CUs. The Base-Z means FBWC without WCC, CTA and FCC.}
	\label{tab6}
	\centering
	\begin{tabular}{c|cl|cc}
		\hline 
		\multirow{2}{*}{Datasets} & \multicolumn{2}{c|}{\multirow{2}{*}{Methods}} & \multicolumn{2}{c}{Results} \\ \cline{4-5} 
		& & & IoU$\uparrow$   & MAE$\downarrow$  \\ \hline
		
		&	A                     & Base-Z $+$ CTA $+$ 1 CU           & 75.39 & 0.115  \\
		&	B                     & Base-Z $+$ CTA $+$ 2 CUs           & 77.51 & 0.108  \\
		HSO &	C                     & Base-Z $+$ CTA $+$ 3 CUs              & 78.42 & 0.010 \\ 
		&	D                     & Base-Z $+$ CTA $+$ 4 CUs            & \textcolor{red}{84.07} & \textcolor{red}{0.079} \\ 
		&	E                     & Base-Z $+$ CTA $+$ 5 CUs           & 80.16 & 0.095  \\ \hline 
		
		&	A                     & Base-Z $+$ CTA $+$ 1 CU          & 76.78 & 0.090  \\
		&	B                     & Base-Z $+$ CTA $+$ 2 CUs          & 78.94 & 0.081  \\
		
		Trans10K-stuff &	C                     & Base-Z $+$ CTA $+$ 3 CUs             & 82.93 & 0.071  \\ 
		&	D                     & Base-Z $+$ CTA $+$ 4 CUs            & \textcolor{red}{86.91} & \textcolor{red}{0.058} \\ 
		&	E                     & Base-Z $+$ CTA $+$ 5 CUs            & 81.47 & 0.077  \\ \hline
		
		&	A                     & Base-Z $+$ CTA $+$ 1 CU        & 86.97 & 0.069  \\
		&	B                     & Base-Z $+$ CTA  $+$ 2 CUs       & 88.76 & 0.062  \\
		
		GDD &	C                     & Base-Z $+$ CTA $+$ 3 CUs             & 89.39 & 0.056  \\ 
		&	D                     & Base-Z $+$ CTA $+$ 4 CUs         & \textcolor{red}{92.79} & \textcolor{red}{0.037}  \\ 
		&	E                     & Base-Z $+$ CTA $+$ 5 CUs        & 85.82 & 0.077  \\ \hline 
		
	\end{tabular}
\end{table}

\begin{table}[t]
	\caption{Quantitative ablation results of FBWC without CTA and FCC on different number of CUs. The Base-Z means FBWC without WCC, CTA and FCC.}
	\label{tab7}
	\centering
	\begin{tabular}{c|cl|cc}
		\hline
		\multirow{2}{*}{Datasets} & \multicolumn{2}{c|}{\multirow{2}{*}{Methods}} & \multicolumn{2}{c}{Results} \\ \cline{4-5} 
		& & & IoU$\uparrow$   & MAE$\downarrow$  \\ \hline
		
		&	A                     & Base-Z  $+$ 1 CU           & 68.92 & 0.130  \\
		&	B                     & Base-Z  $+$ 2 CUs           & 71.33 & 0.118  \\
		HSO &	C                     & Base-Z  $+$ 3 CUs              & 70.69 & 0.121 \\ 
		&	D                     & Base-Z $+$ 4 CUs            & \textcolor{red}{77.41} & \textcolor{red}{0.093} \\ 
		&	E                     & Base-Z $+$ 5 CUs           & 71.84 & 0.116  \\ \hline 
		
		&	A                     & Base-Z $+$ 1 CU          & 72.71 & 0.150  \\
		&	B                     & Base-Z  $+$ 2 CUs          & 74.25 & 0.120  \\
		
		Trans10K-stuff &	C                     & Base-Z  $+$ 3 CUs             & 78.43 & 0.092  \\ 
		&	D                     & Base-Z $+$ 4 CUs            & \textcolor{red}{79.92} & \textcolor{red}{0.090} \\ 
		&	E                     & Base-Z $+$ 5 CUs            & 74.81 & 0.100  \\ \hline
		
		&	A                     & Base-Z $+$ 1 CU        & 82.87 & 0.087  \\
		&	B                     & Base-Z $+$ 2 CUs       & 83.16 & 0.085  \\
		
		GDD &	C                     & Base-Z $+$ 3 CUs             & 84.92 & 0.082  \\ 
		&	D                     & Base-Z  $+$ 4 CUs         & \textcolor{red}{86.05} & \textcolor{red}{0.073}  \\ 
		&	E                     & Base-Z  $+$ 5 CUs        & 81.93 & 0.090  \\ \hline 
		
	\end{tabular}
\end{table}

In addition, in order to further verify the effectiveness of the proposed module, we conducted more in-depth ablation experiments on CTA and FCC. Specifically, on three datasets, we conducted three sets of ablation experiments with different CU numbers: FBWC without CTA, FBWC without FCC, and without both CTA and FCC. The experiment aims to further explore the effects of CTA and FCC on the segmentation accuracy of the network under different CU numbers.

It can be observed from the data shown in Table \ref{tab5}, under GDD dataset, the network segmentation performance of the FBWC without CTA experimental group becomes better with the increase of the number of CUs, and the best advantage occurs when the number of CUs is 4. When the number of CUs is 5, the network segmentation performance decreases to 86.39\% IoU and 0.073 MAE. By horizontal contrast among Table \ref{tab5}, Table \ref{tab6}, and Table  \ref{tab7}, when the number of CUs is constant, the experimental group without both CTA and FCC in FBWC performs the most poorly compared with the other two groups. To be more specific, we observed that when the number of CUs was 3 in the HSO dataset, the network segmentation results without CTA and FCC are 70.69\% IoU and 0.121 MAE, which are worse than the others with the same number of CUs. At the same time, under the experimental settings above mentioned, the segmentation performance results shown in Table \ref{tab5} and Table \ref{tab7} are lower than those shown in Table \ref{tab6}. This reflects the importance and effectiveness of the FCC in the FBWC.

The experimental phenomena mentioned above is also reflected among three datasets with different numbers of CUs. When the number of CUs is 4, the overall network can have better segmentation performance.

\begin{figure}[t]
	\begin{center}
		\includegraphics[width=1\linewidth]{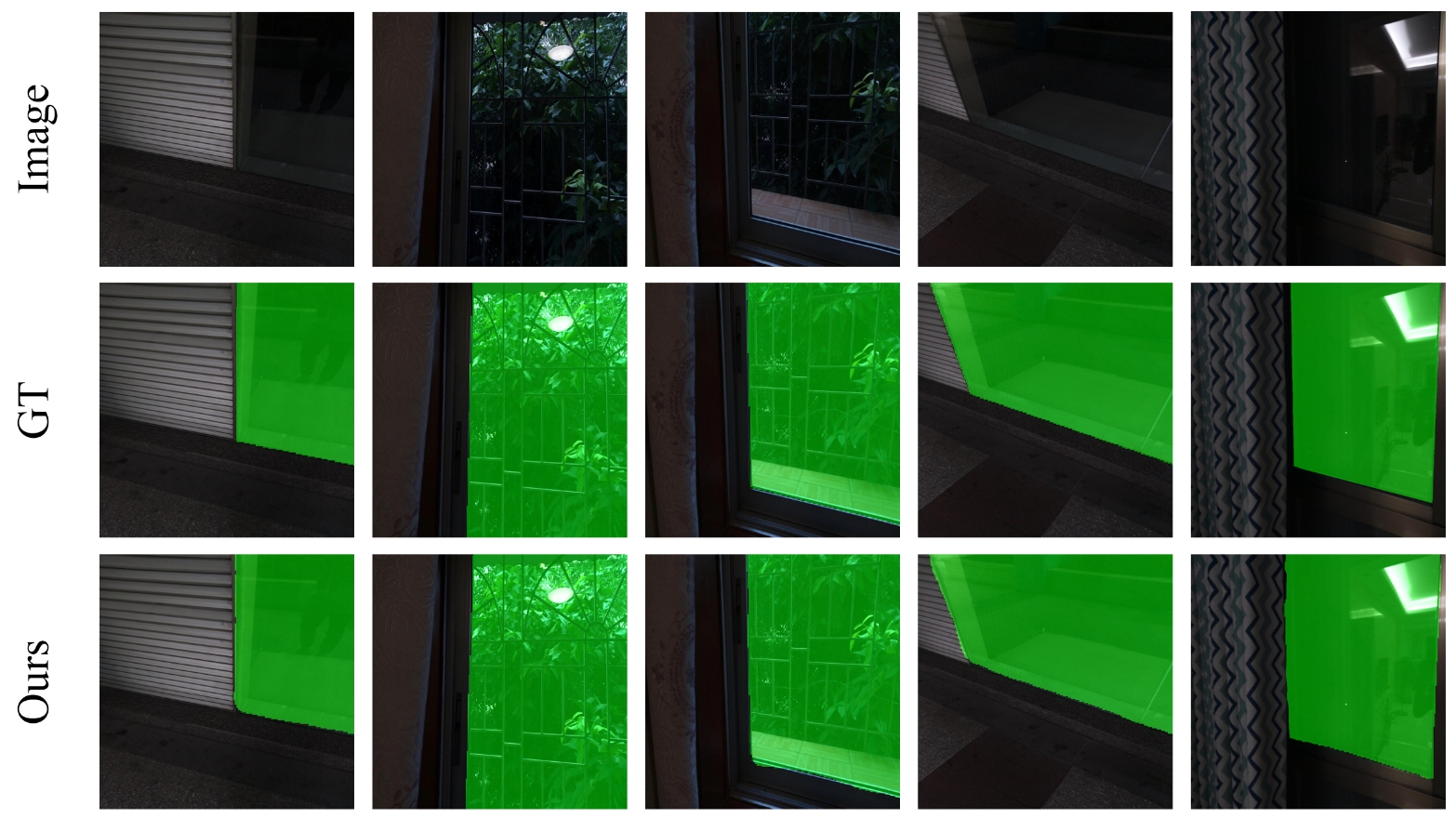}
	\end{center}
	\vspace{-12pt}
	\caption{Visual results of FBWC on night scene .}
	\label{fig11}
\end{figure}

\begin{figure}[t]
	\begin{center}
		\includegraphics[width=1\linewidth]{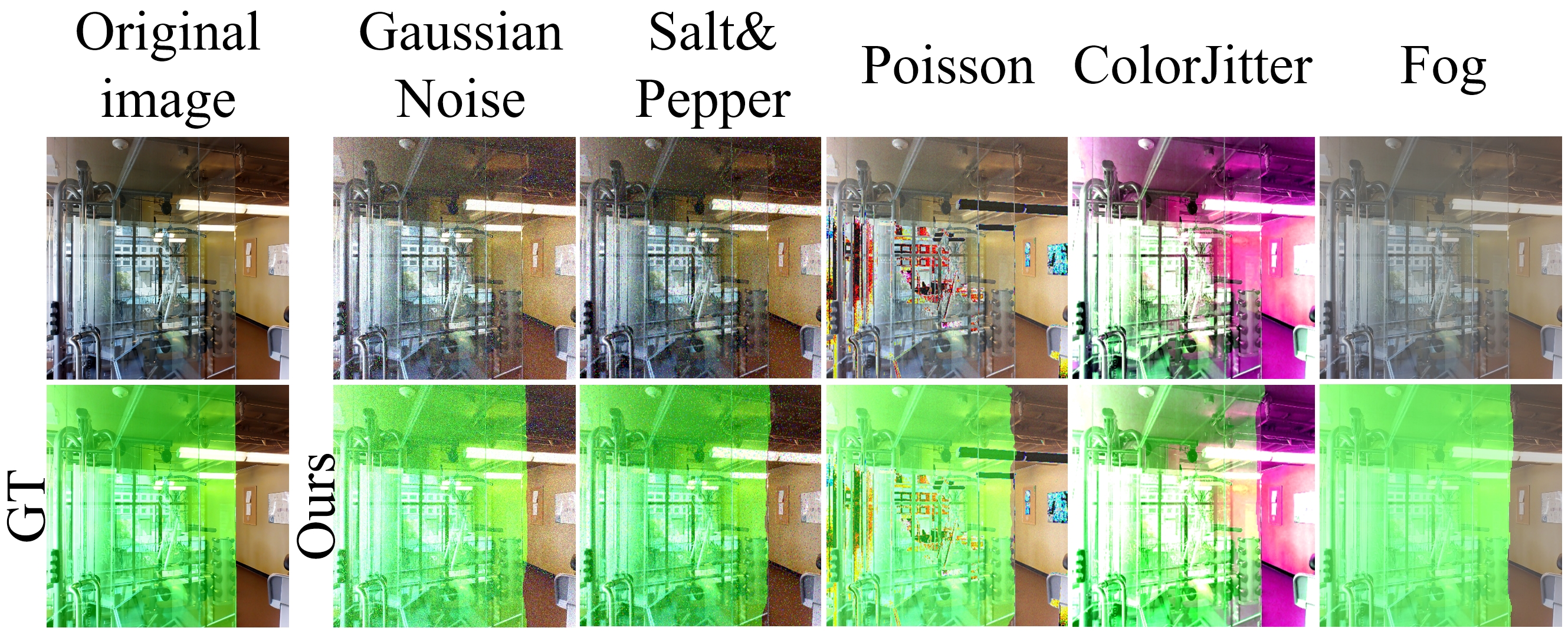}
	\end{center}
	\vspace{-12pt}
	\caption{Visual results of FBWC on noise condition.}
	\label{fig12}
\end{figure}

\begin{figure}[t]
	\begin{center}
		\includegraphics[width=1\linewidth]{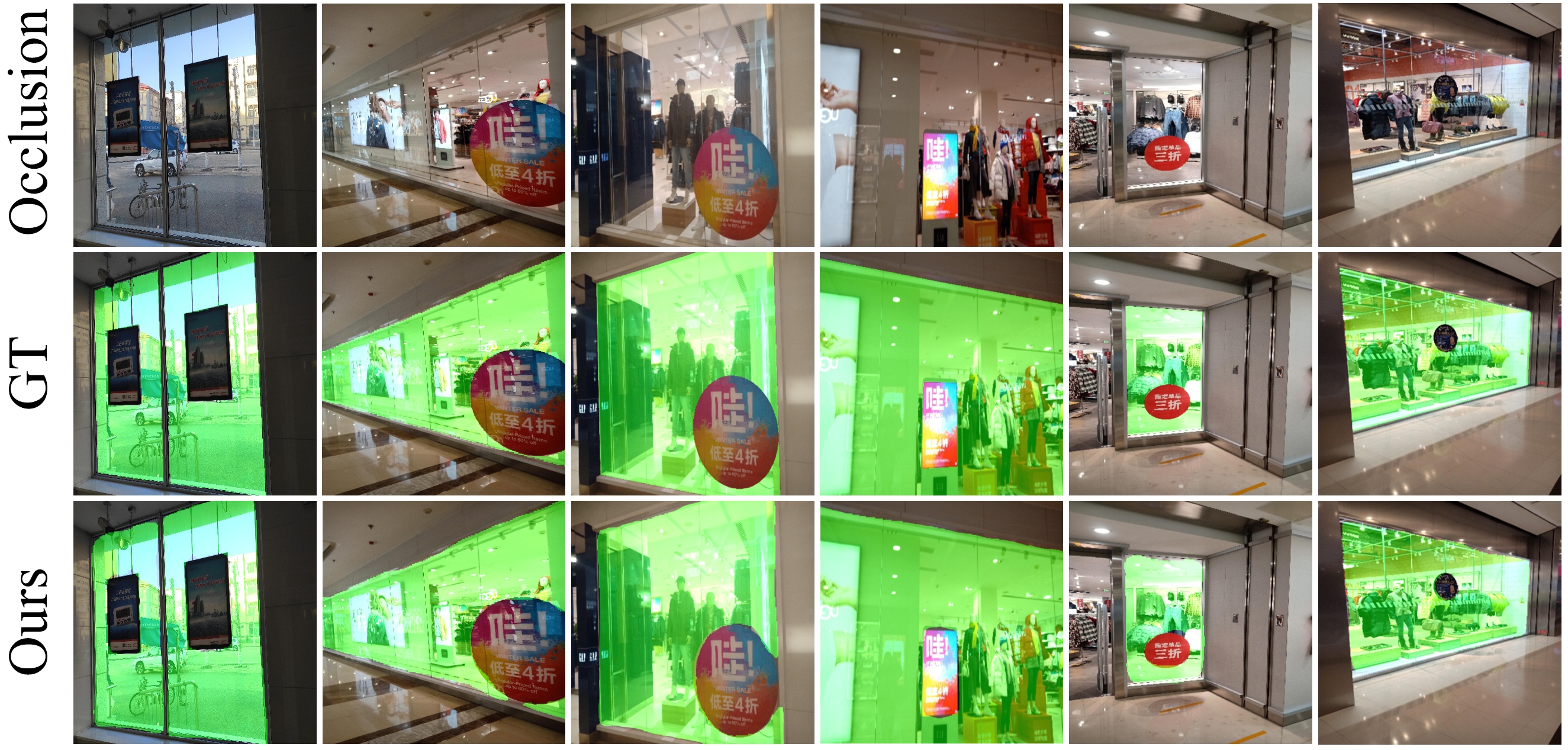}
	\end{center}
	\vspace{-12pt}
	\caption{Visual results of FBWC on occlusion condition. }
	\label{fig13}
\end{figure}

\subsection{Discussion and Analysis}
\subsubsection{Night Scenes}
In night scenes, due to poor lighting conditions, the boundary information in the image becomes blurry, and the semantic features of large glass areas are not significant. To address this issue, the proposed FBWC could effectively segment glass regions as shown in Figure \ref{fig11}. 

For the problem of boundary blurring under low light conditions, the WCC and continuous strong boundary constraints can effectively capture boundary features while preventing the dispersion of fine-grained boundary semantic information. Meanwhile, CTA enhances the consistency of glass regions through attention mechanisms. Finally, the boundary information and coarse-grained semantic features are further highlighted and effectively integrated through FCC. Based on the above content, FBWC can effectively cope with semantic segmentation tasks of glass images in night scenes. In addition, the datasets utilized in this work contain real-world glass images under low light conditions, providing sample diversity and further enhancing the model's generalization ability.

\subsubsection{Non-ideal Conditions}
Similarly, when dealing with non-ideal conditions such as noise and occlusion in the glass semantic segmentation task, boundary and glass region features are easily affected by noise interference and are difficult to extract. The segmentation results of the proposed FBWC under Gaussian noise, Salt\&Pepper noise, Poisson noise, ColorJitter, and Foggy noise are shown in Figure \ref{fig12}, and the segmentation results under occlusion conditions are shown in Figure \ref{fig13}.

In order to effectively address noise and occlusion issues, FBWC applies WCC to avoid the over-capturing of large glass areas, which is beneficial for maintaining the integrity of glass areas under occlusion, light spots, and noise interference. CTA is further designed to achieve effective segmentation of non-occluded areas. Simultaneously, through multi-level boundary loss constraints and effective boundary enhancement of FFT, the boundary information in shallow structures has been significantly optimized, further improving the anti-interference ability and robustness of FBWC.

\subsubsection{Complexity and Efficiency}
We have indicated the parameter count for each module in Figure \ref{fig2}. It is worth mentioning that the parameters of the whole FBWC are 48.7M and the backbone WCC we designed is 38.38M. Regarding the backbone, WCC reduces the parameters of 47.62M, 99.62M and 5.62M compared to ViT-Base\cite{dosovitskiy2020image}, VGG16\cite{simonyan2014very} and ResNeXt101\cite{mahajan2018exploring}, whose are used in the popular methods of glass segmentation. In addition, the parameters of the proposed FCC and SegHead are 2.37M, which does not significantly increase the parameters of the FBWC. This optimization not only reduces the computational complexity of the model but also improves the training and inference efficiency of the model. By reducing the number of parameters, WCC reduces the storage and computational burden of the model while maintaining high performance.

\section{Conclusion and Future Work}
In this paper, high-precision glass segmentation is realized by strengthening the glass boundary and avoiding over-capturing semantic information in deep structure caused by the glass property of reflectivity and light transmission. We connect the designed CUs horizontally to appropriately capture large area semantic features with strong boundary constraints. Besides, CTA is embedded to keep the feature region consistent. The FCC is then utilized for the different features, which can be flexibly adjusted in a learnable way and integrated effectively. Extensive experimental results show that our proposed FBWC produces comprehensive SOTA performance on the three different public glass segmentation datasets. In the future, we will explore the application of shallow network architecture in more scenarios.

\begin{figure}[t]
	\begin{center}
		\includegraphics[width=1\linewidth]{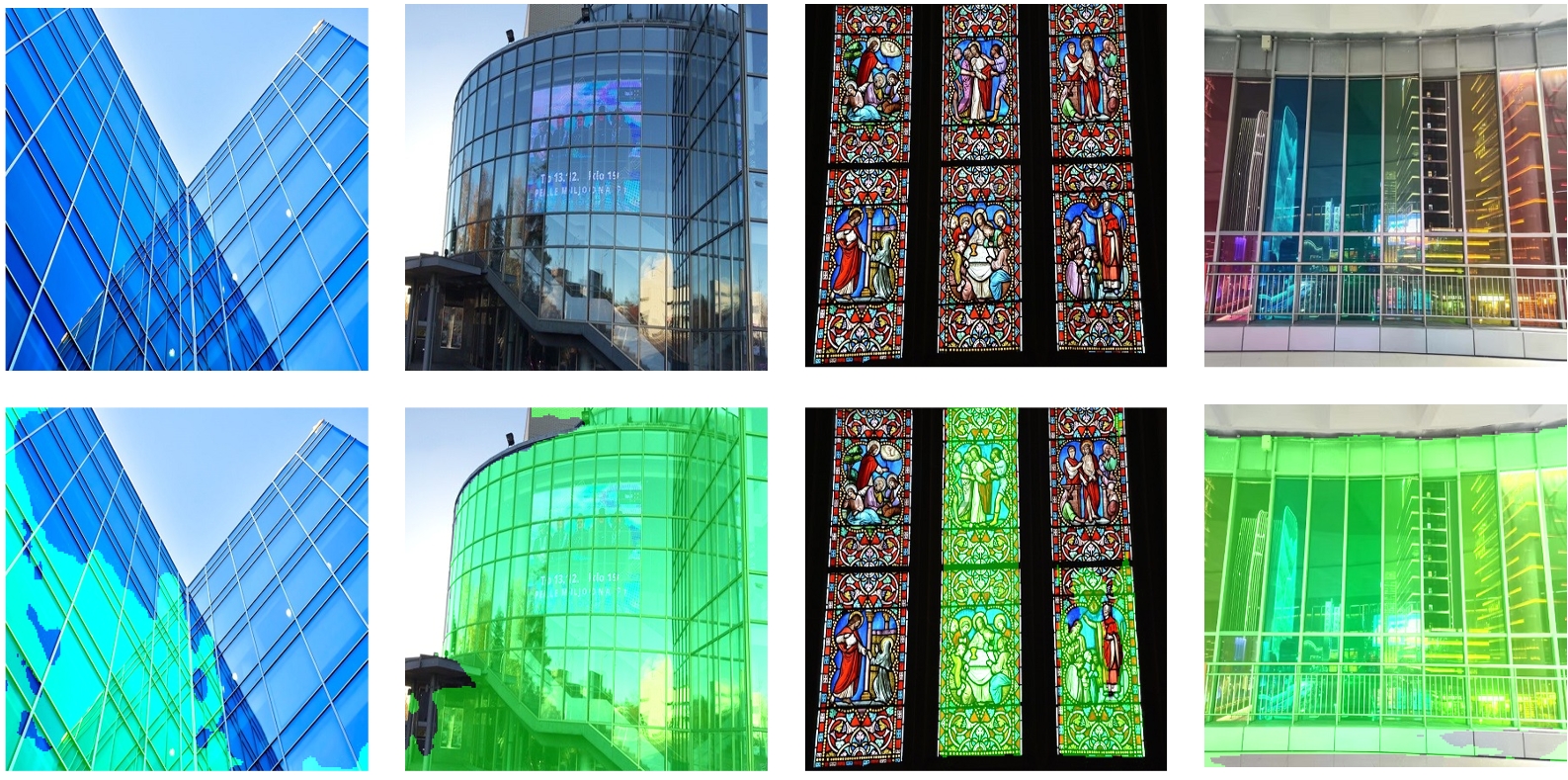}
	\end{center}
	\vspace{-12pt}
	\caption{Visualization of poor results of glass group and stained glass segmentation based on our works.}
	\label{fig14}
\end{figure}

At the same time, we also objectively noted that the proposed method still has some limitations. When segmenting specific objects, as shown in Figure \ref{fig14}, for buildings covered by large glass groups, the network cannot effectively explore such boundaries to achieve fine and complete segmentation. When the stained glass with decorative function is segmented, the results of the method are not ideal. 

After analysis, we believe that the possible reasons are as follows: Different from conventional glass, colorful decorative glass to improve the decoration, the color fullness of the glass itself and the complexity of the pattern will be greatly increased, which leads to an increase in the difficulty of retaining the general characteristics of glass, such as transparency, reflective performance, and so on. From the perspective of semantic information, the semantic features of such glass are more difficult to distinguish from context. Furthermore, decorative colored glass with high texture and color richness poses challenges for extracting boundary information. And the proposed boundary constraint method requires the support of real data, thus placing higher demands on the quality of data annotation, and is affected by the subjectivity and accuracy of manual annotation.

Therefore, in future work, it is necessary to introduce more image data containing complex boundaries or colored glass, so that the model can better complete diverse glass image segmentation tasks. Meanwhile, we will explore semi-supervised or unsupervised learning methods to reduce reliance on high-quality annotated data.

\section*{Acknowledgment}

This research was partly supported by the Sichuan Science and Technology Program (2024NSFJQ0035), the Talents by Sichuan provincial Party Committee Organization Department, and Chengdu - Chinese Academy of Sciences Science and Technology Cooperation Fund Project (Major Scientific and Technological Innovation Projects).

\bibliographystyle{IEEEtran}

\bibliography{dmain}

\vspace{-1.5cm}
\begin{IEEEbiography}[{\includegraphics[width=1in,height=1.25in,clip,keepaspectratio]{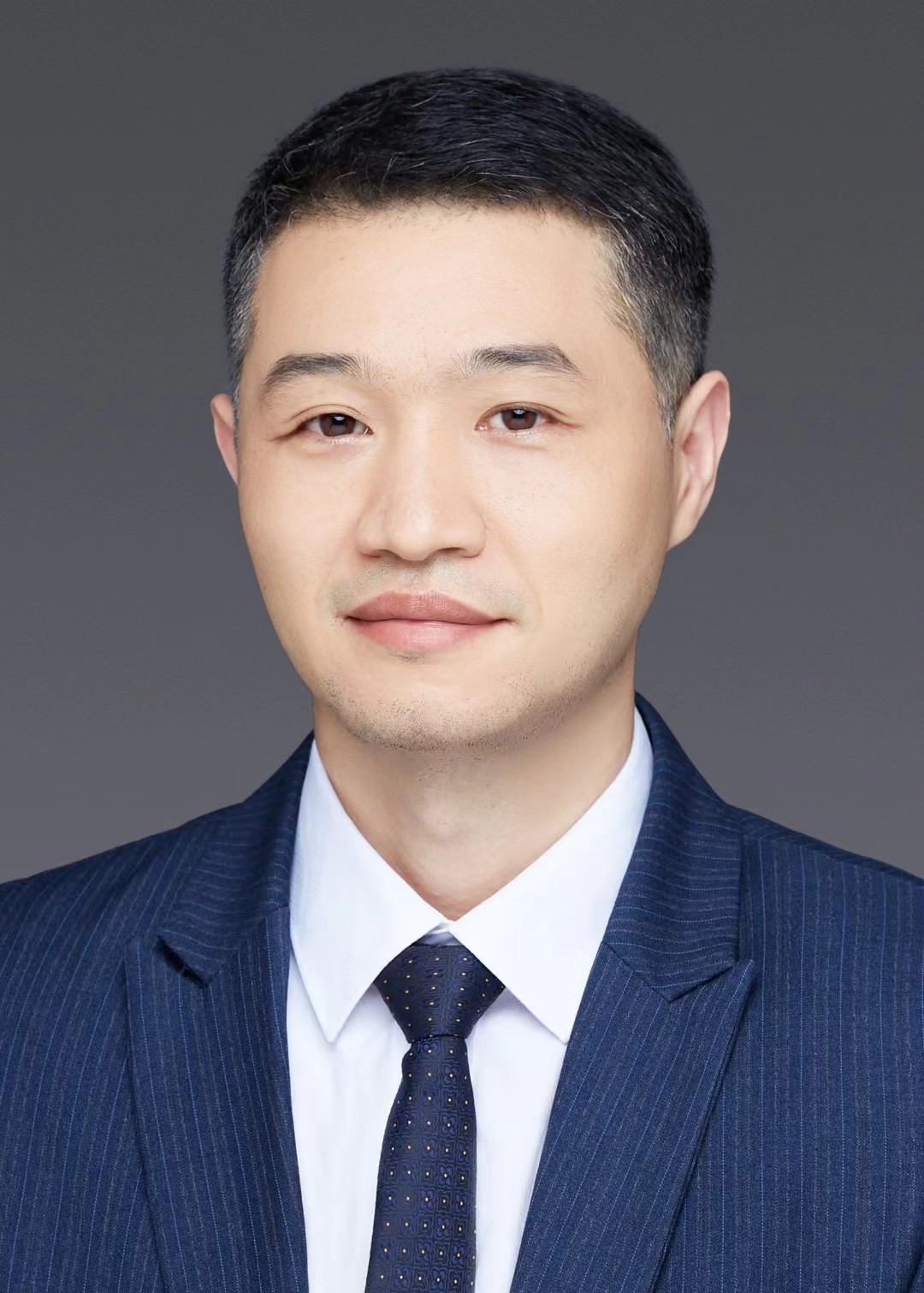}}]{Xiaolin Qin (Senior Member, IEEE)}
	is a Professor in Chengdu Institute of Computer Applications, Chinese Academy of Sciences and University of Chinese Academy of Sciences, China. Qin received his Ph.D. degree in July 2011. From May 2014 to June 2015, he was a postdoctoral fellow at Department of Computer and Information Science, Linköping University, Sweden. His research interests include automated reasoning and algebraic vision.\end{IEEEbiography}

\vspace{-1.5cm}
\begin{IEEEbiography}[{\includegraphics[width=1in,height=1.25in,clip,keepaspectratio]{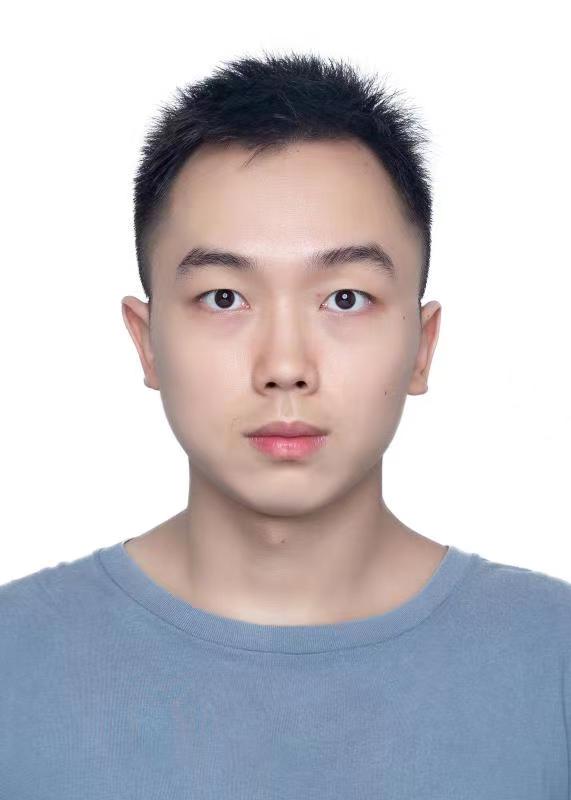}}]{Jiacen Liu}
	is a research assistant  of  Chengdu Institute of Computer Applications at Chinese Academy of Sciences, China. Liu received his Postgraduate degree in June 2024 from Kunming University of Science and Technology. His research interests include image processing and algebraic vision.\end{IEEEbiography}

\vspace{-1.5cm}
\begin{IEEEbiography}[{\includegraphics[width=1in,height=1.25in,clip,keepaspectratio]{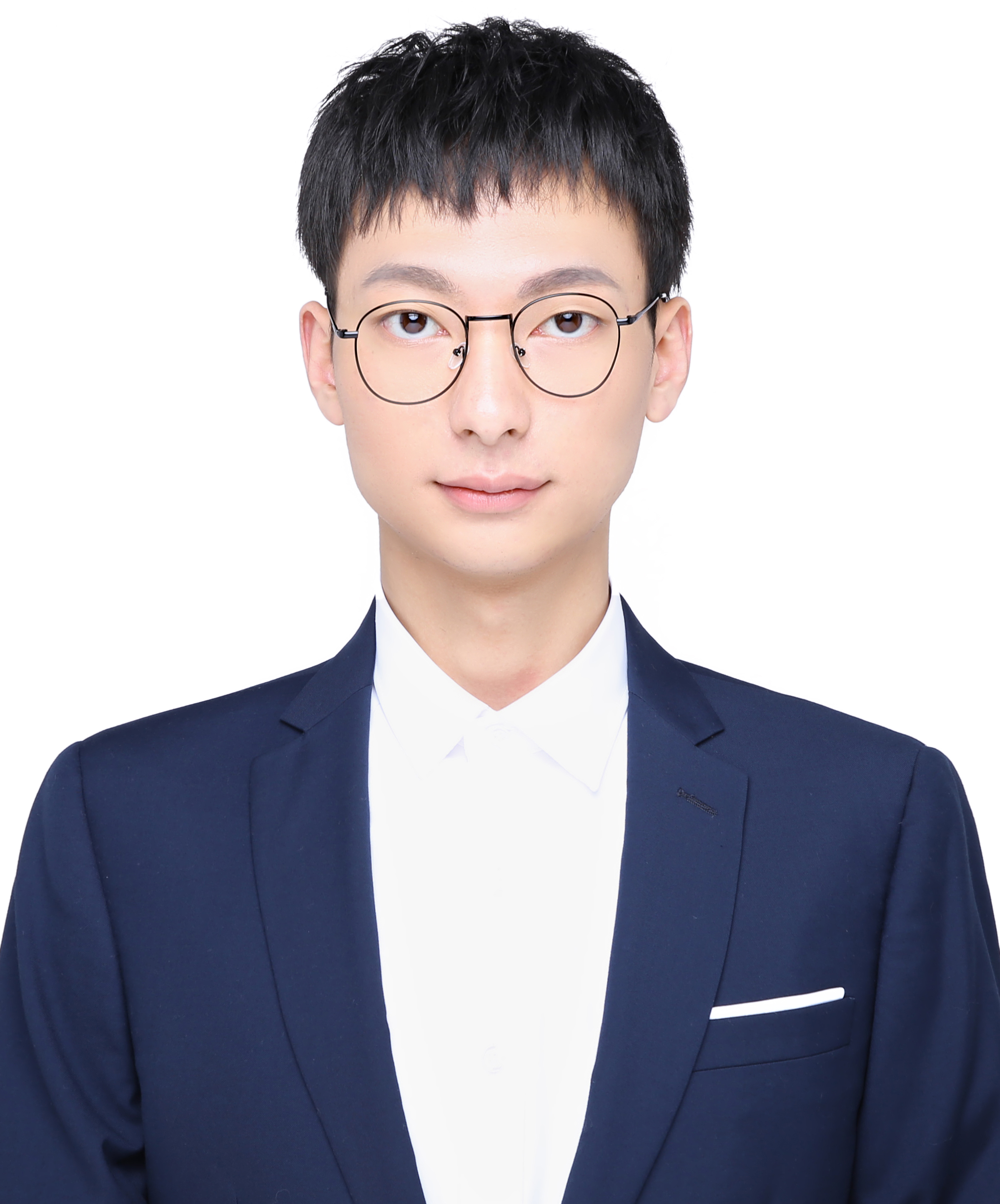}}]{Qianlei Wang}
	received the M.S. degree in Computer Application from Civil Aviation Flight University of China, China, in July 2022. He is currently working toward the Ph.D. degree with the School of Computer Science and Technology at University of Chinese Academy of Sciences, China. His research interests include SLAM and algebraic vision.\end{IEEEbiography}

\vspace{-1.5cm}
\begin{IEEEbiography}[{\includegraphics[width=1in,height=1.25in,clip,keepaspectratio]{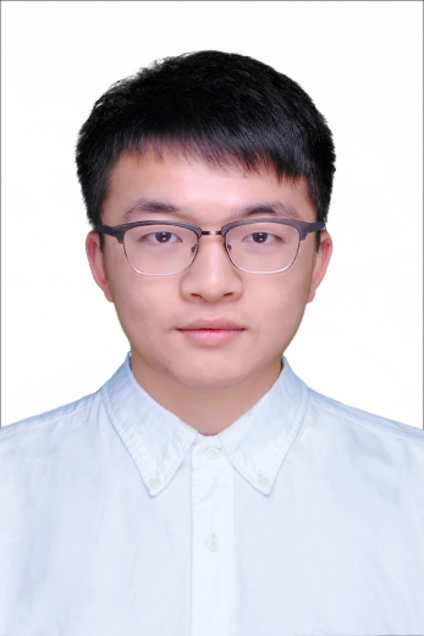}}]{Shaolin Zhang}
	is currently working toward the Ph.D. degree with the School of Computer Science and Technology at University of Chinese Academy of Sciences, China. He received his PostGraduate degree in July 2023 from Chongqing University of Science and Technology, China. His research interests include image processing and monocular depth estimation.\end{IEEEbiography}

\vspace{-1.5cm}
\begin{IEEEbiography}[{\includegraphics[width=1in,height=1.25in,clip,keepaspectratio]{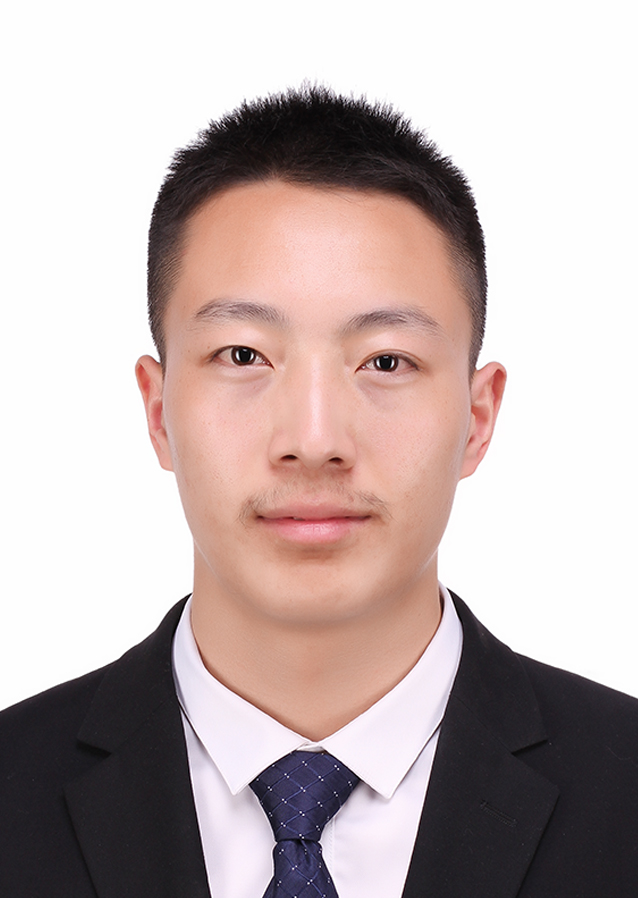}}]{Fei Zhu} received the B.E. degree in mechanical engineering from Tsinghua University, China, in 2018, and the Ph.D. degree in pattern recognition and intelligent systems from the Institute of Automation, Chinese Academy of Sciences, China, in 2023. He is now a postdoctoral fellow at the Centre for Artificial Intelligence and Robotics, Hong Kong Institute of Science \& Innovation, Chinese Academy of Sciences. His research interests include open-world and trustworthy machine learning.
\end{IEEEbiography}

\vspace{-1.5cm}
\begin{IEEEbiography}[{\includegraphics[width=1in,height=1.25in,clip,keepaspectratio]{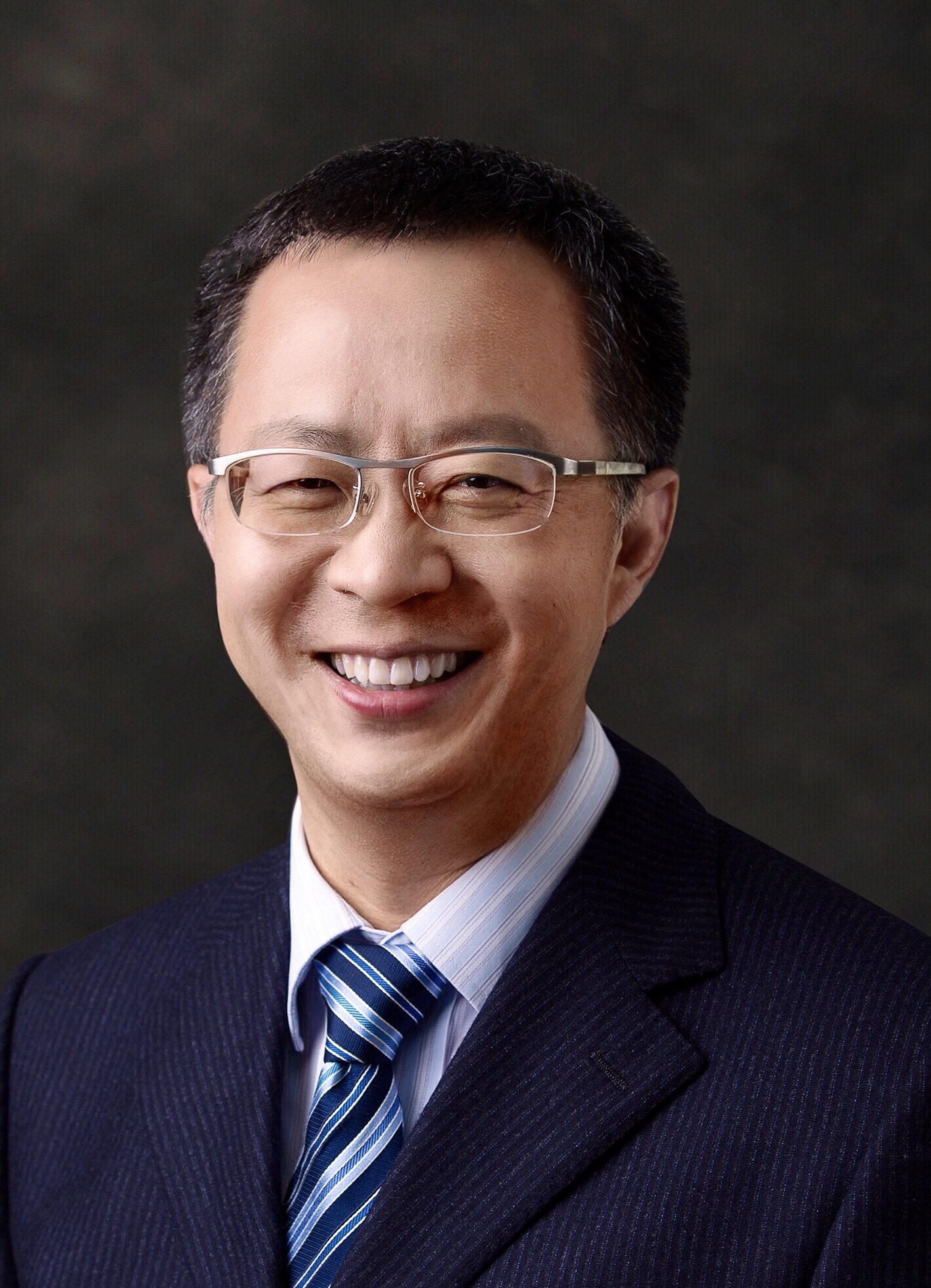}}]{Zhang Yi (Fellow, IEEE)}
	received the Ph.D. degree in mathematics from the Institute of Mathematics, Chinese Academy of Sciences, China, in 1994. He is currently a Professor with the College of Computer Science, Sichuan University, China. He coauthored three books entitled Convergence Analysis of Recurrent Neural Networks (Kluwer, 2004), Neural Networks: Computational Models and Applications (Springer, 2007), and Subspace Learning of Neural Networks (CRC, 2010). His research interests include neural networks and intelligent medicine. He was an Associate Editor for IEEE TRANSACTIONS ON CYBERNETICS. From 2009 to 2012, he was an Associate Editor for IEEE TRANSACTIONS ON NEURAL NETWORKS AND LEARNING SYSTEMS. He is a Foreign Member of Russian Academy of Engineering.\end{IEEEbiography}

\end{document}